\documentclass[runningheads]{llncs}

\usepackage{eccv}

\usepackage{eccvabbrv}

\usepackage{graphicx}
\usepackage{booktabs}
\usepackage[table]{xcolor}
\usepackage[accsupp]{axessibility}  

\usepackage{hyperref}

\usepackage{orcidlink}

\usepackage[percent]{overpic} 
\usepackage{booktabs}
\usepackage{multirow}
\usepackage{makecell}
\usepackage{array}
\usepackage{graphicx}

\definecolor{gold}{RGB}{198,224,180}
\definecolor{silver}{RGB}{235,245,230}
\definecolor{bronze}{RGB}{245,249,242}

\newcommand{\best}[1]{\textbf{#1}}
\newcommand{\second}[1]{\underline{#1}}
\newcommand{\third}[1]{{#1}}

\usepackage{marvosym} 
\newcommand{\corrauth}{\textsuperscript{\Letter}}

\usepackage[table]{xcolor}
\usepackage{colortbl}
\usepackage{booktabs}
\usepackage{array}
\definecolor{oursrow}{RGB}{245,245,245}
\usepackage{threeparttable}

\usepackage{stfloats}
\usepackage{cuted}
\usepackage{capt-of}

\usepackage{needspace} 

\usepackage[normalem]{ulem}

\begin{document}

\title{Any to Full: Prompting Depth Anything for Depth Completion in One Stage} 

\titlerunning{Any to Full}


\author{Zhiyuan Zhou\inst{1} \and
Ruofeng Liu\inst{2} \corrauth  \and
Taichi Liu\inst{1}\and
Weijian Zuo\inst{3}\and
Shanshan Wang \inst{1}\and
Zhiqing Hong\inst{4}\and
Desheng Zhang\inst{1}}

\authorrunning{Z.~Zhou et al.}



\institute{
    Rutgers University, USA \ \email{\{zhiyuan.z,taichi.liu,shanshan.wang\}@rutgers.edu, desheng@cs.rutgers.edu} \and
    Michigan State University, USA \ \email{liuruofe@msu.edu} \and
    JD Logistics, China \ \email{zuoweijian1@jd.com} \and
    HKUST (Guangzhou), China \ \email{zhiqinghong@hkust-gz.edu.cn}
}

\maketitle

\begingroup
\renewcommand{\thefootnote}
{\textrm{\Letter}} \footnotetext[1]
{Corresponding author.} \endgroup

\begin{abstract}

Accurate, dense depth estimation is crucial for robotic perception, but commodity sensors often yield sparse or incomplete measurements due to hardware limitations. 
Existing RGBD-fused depth completion methods learn priors jointly conditioned on training RGB distribution and specific depth patterns, limiting domain generalization and robustness to various depth patterns. Recent efforts leverage monocular depth estimation (MDE) models to introduce domain-general geometric priors, but current two-stage integration strategies relying on explicit relative-to-metric alignment incur additional computation and introduce structured distortions.
To this end, we present Any2Full, a one-stage, domain-general, and pattern-agnostic framework that reformulates completion as a scale-prompting adaptation of a pretrained MDE model. 
To address varying depth sparsity levels and irregular spatial distributions, we design a Scale-Aware Prompt Encoder. It distills scale cues from sparse inputs into unified scale prompts, guiding the MDE model toward globally scale-consistent predictions while preserving its geometric priors. 
Extensive experiments demonstrate that Any2Full achieves superior robustness and efficiency. It outperforms OMNI-DC by 32.2\% in average AbsREL and delivers a 1.4× speedup over PriorDA with the same MDE backbone, establishing a new paradigm for universal depth completion. Codes and checkpoints are available at \url{https://github.com/zhiyuandaily/Any2Full}.


\end{abstract}    

\section{Introduction}
\label{sec:intro}
Accurate and fine-grained depth information is essential for robotic perception, enabling navigation~\cite{tang2022perception,maier2012real}, manipulation~\cite{mahler2019learning,du2021vision,tan2024attention}, and scene understanding~\cite{armeni2017joint,chen2019towards,miao2023occdepth}. However, commodity depth sensors (e.g., LiDAR~\cite{geiger2012we}, ToF~\cite{foix2011lock}, or structured light cameras~\cite{schonberger2016structure}) often yield sparse or incomplete depth maps due to limitations in resolution, range, and light interaction, such as reflection or absorption~\cite{wang2024depthholev2}. Consequently, depth completion has emerged as a fundamental task that aims to recover dense metric depth maps from raw depth measurements and their corresponding RGB images.

\begin{figure}[t]
    \centering
    \begin{overpic}[width=1\linewidth]{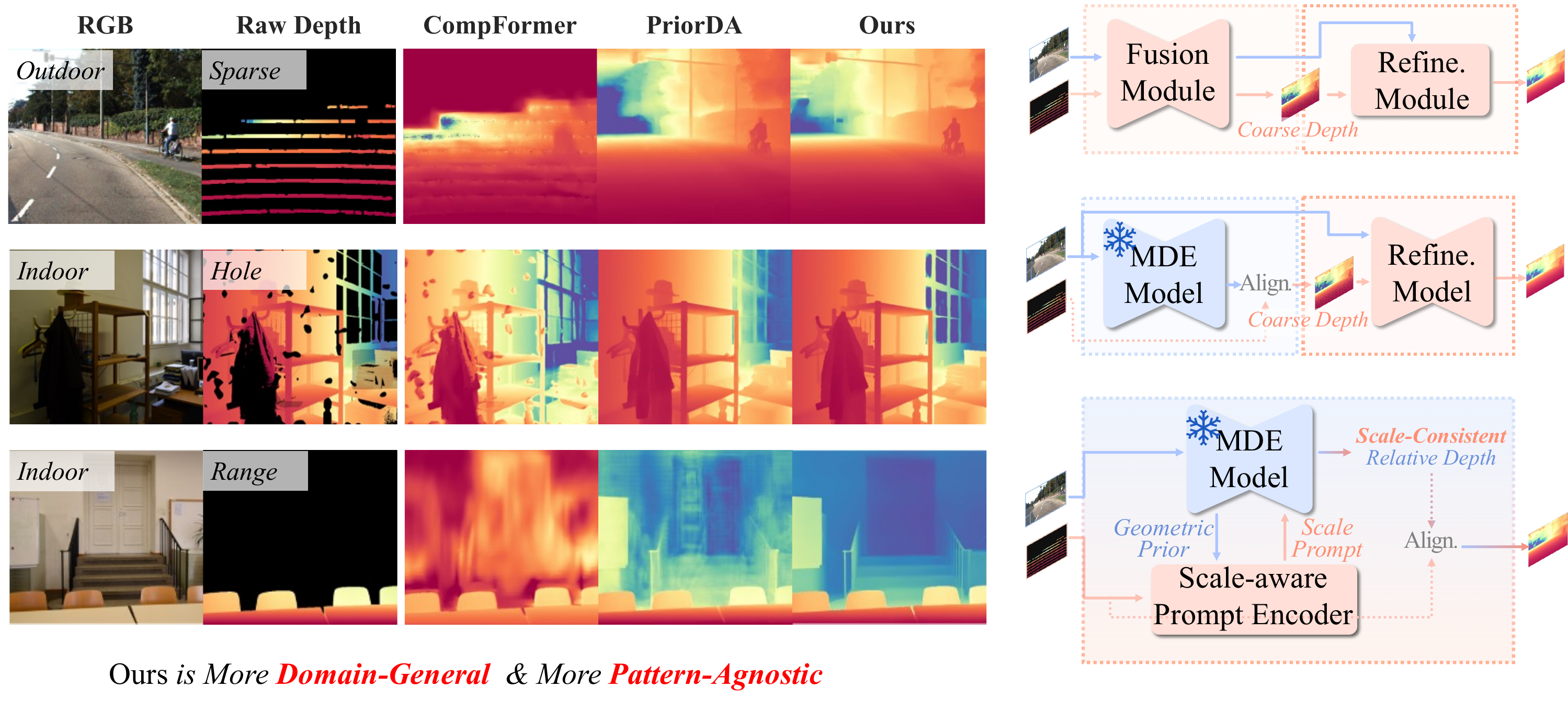}

       \put(63.5,33.5){\scalebox{0.85}[1]{\scriptsize
            \textbf{(a)} 2-stage RGBD fused 
            (\textit{e.g., CompFormer})
        }}

        \put(63.5,21){\scalebox{0.85}[1]{\scriptsize
            \textbf{(b)} 2-stage MDE integrated 
            (\textit{e.g., PriorDA})
        }}

        \put(63.5,1){\scriptsize
            \textbf{(c)} 1-stage MDE integrated (ours)
        }
    \end{overpic}
    \caption{Depth completion recovers dense depth maps from raw measurements and RGB guidance.
(a) Traditional two-stage methods predict coarse depth to bridge the gap between sparse input and dense output.
(b) Recent approaches integrate monocular depth estimation (MDE) to generate relative depth and explicitly align it with sparse depth, disrupting MDE’s geometric priors.
(c) Our framework employs a lightweight prompting mechanism that tightly integrates MDE’s geometric priors, achieving domain-general and pattern-agnostic depth completion in one stage.}
\vspace{-5pt}
    \label{fig:depth_stages}
\end{figure}

Most existing depth completion methods, as illustrated in~\cref{fig:depth_stages}(a), learn geometric priors from RGB images to guide dense depth prediction from sparse inputs~\cite{cheng2018depthcspn,park2020nonlocalspn,tang2020learning,rho2022guideformer,tang2024bispn,zhang2023completionformer,jun2024masked}.
However, as shown in~\cref{fig:depth_stages}(left), these RGBD fused frameworks, such as CompFormer~\cite{zhang2023completionformer}, learn priors jointly conditioned on RGB distribution and specific depth patterns observed during training, leading to two commonly observed limitations: 
(1) \textbf{Domain Specificity}, where performance degrades under visual domain shifts such as lighting, texture, or scene variations. 
(2) \textbf{Depth Pattern Sensitivity}, where performance degrades with changing raw depth patterns caused by heterogeneous depth sensors, such as variations in density, missing regions, or limitations in sensing range. 

Recent monocular depth estimation (MDE) models~\cite{yang2024depthanythingv2,ke2024marigold} show remarkable domain generalization, inspiring recent efforts~\cite{viola2025marigolddc,wang2025priorda,cheng2025monster} to leverage their geometric priors for depth completion.
Nevertheless, current integration strategies remain suboptimal. Test-time adaptation methods~\cite{viola2025marigolddc} are too computationally expensive for real-time use, while direct feed-forward methods~\cite{park2024depthprompt,wang2025priorda} typically follow a traditional two-stage design (\cref{fig:depth_stages}(a), (b)): an initial stage predicts coarse depth to bridge the gap between sparse to dense output, followed by a refinement stage to recover structural details. In these methods, the RGBD fusion in the first stage is replaced by explicit alignment between relative depth from MDE and sparse metric inputs to produce a coarse metric map.

However, despite effectively addressing the input depth sparsity, this two-stage paradigm not only incurs extra computational overhead but also suffers from MDE's inherent scale inconsistency. 
As analyzed in Supp.~\ref{supsec:scale}, relative depth predictions often exhibit non-linear distortions that require spatially varying scale factors to align with metric ground truth~\cite{hyoseok2025zero}. 
Explicitly aligning such inconsistent depth maps introduces structured distortions and depth-pattern-specific biases~\cite{wang2025tacodepth}, ultimately resulting in noticeable artifacts on unseen depth patterns (i.e. Range), as illustrated by the PriorDA results in~\cref{fig:depth_stages} (left).
These observations raise a fundamental question: \textit{How can we bypass generation of intermediate coarse depth and instead seamlessly integrate MDE's geometric priors in one stage, achieving domain-general and pattern-agnostic depth completion?}

To answer this question, we propose Any2Full (\cref{fig:depth_stages} (c)), a one-stage framework that reformulates depth completion as a scale-prompting adaptation of MDE. It distills scale cues (i.e., inter-point scale ratios) from sparse inputs to prompt the pretrained MDE model toward scale-consistent predictions while preserving its robust domain generalization. Here, ``\textbf{scale-consistent}'' means that the predicted relative depth can be aligned to metric depth using a global scale and bias across the scene.
However, constructing effective scale prompts from sparse depth is challenging, as sparse inputs not only exhibit varying sparsity levels but also follow irregular spatial distributions (e.g., random missing regions). 
Such pattern heterogeneity leads to unstable training and often triggers pattern-specific overfitting.
To address these issues, we reconstruct the structural context of sparse depth input into unified scale prompt representation under the guidance of MDE’s geometric priors, preserving essential scale cues while decoupling the scale-prompting process from specific depth patterns.

Specifically, we design a Scale-Aware Prompt Encoder with two hierarchical modules.
First, the Local Enrichment module couples scale cues from sparse depth with MDE’s dense geometric context, producing local scale-aware features that are robust to varying sparsity levels.
Second, the Global Propagation module further propagates these features across the scene through MDE geometry-aware attention, addressing irregular spatial distributions and establishing globally consistent scale-aware features.
Finally, the resulting features form a unified scale prompt that modulates MDE features, enabling one-stage, domain-general, and pattern-agnostic depth completion.

In summary, this work has the following contributions: 
\begin{itemize}
    \item We introduce \textbf{Any2Full}, a one-stage framework that completes \textit{any} incomplete depth input into a \textit{full} dense depth map. By reformulating depth completion as a scale-prompting adaptation of monocular depth estimation (MDE), it unleashes the power of pretrained MDE models for domain-general and pattern-agnostic depth completion.

    \item We design a Scale-Aware Prompt Encoder that hierarchically transforms sparse and irregular scale cues into globally consistent, pattern-invariant scale-aware features under MDE geometric guidance, enabling robust scale prompting with minimal additional computation.  

    \item Extensive experiments across diverse domains and depth patterns demonstrate that Any2Full outperforms SOTA (OMNI-DC) by 32.2\% in average AbsREL. Compared to the competitive PriorDA using the same MDE backbone, Any2Full achieves a 1.4$\times$ speedup with superior precision. These results highlight the efficiency of our one-stage scale prompting framework, while its real-world deployment in robotic warehouse grasping further demonstrates its practical value.
\end{itemize}

\section{Related Work}

\label{sec:relatedwork}


\subsection{Monocular Depth Estimation}

Monocular depth estimation (MDE)~\cite{eigen2014depth,ranftl2020towards} aims to infer the 3D scene structure from a single RGB image. It is an ill-posed inverse problem due to inherent scale ambiguity, which means that many distinct 3D scenes can produce identical image projections. Nonetheless, relative depth relationships can be inferred from visual cues such as texture gradients, object sizes, and occlusions~\cite{arampatzakis2023monocularrevier}.

Inspired by the success of foundation models in natural language processing~\cite{devlin2019bert,brown2020language}, MDE has evolved toward strong zero-shot generalization~\cite{ranftl2020towards,yang2024depthanythingv1,ke2024marigold,moon2024ground,bhat2023zoedepth,bochkovskii2024depth,gui2025depthfm,wang2025moge,wang2026moge}.
Early efforts like MiDaS~\cite{ranftl2020towards} leveraged mixed-dataset training, while Depth Anything~\cite{yang2024depthanythingv2} utilizes massive unlabeled images to learn robust relative geometric priors. In parallel, Marigold~\cite{ke2024marigold} converts pre-trained latent diffusion models into conditional depth estimators to achieve superior generalization.
While the above MDE models are limited to relative depth prediction due to scale ambiguity, some recent methods~\cite{bhat2023zoedepth,yin2023metric3d,zhu2024scaledepth} like UniDepth~\cite{piccinelli2024unidepth,piccinelli2025unidepthv2} and  WorDepth~\cite{zeng2024wordepth} incorporate camera or object priors to enable monocular metric depth estimation.
Nevertheless, the scale ambiguity remains fundamentally unsolved in the absence of reliable metric measurements.



\subsection{Depth Completion}
Depth completion~\cite{ma2018sparse} aims to recover dense metric depth from sparse inputs and RGB images. Conventional methods adopt a two-stage pipeline~\cite{cheng2018depthcspn,cheng2019learning,cheng2020cspn++,park2020nonlocalspn,lin2022dynamic,liu2022graphcspn,tang2024bispn,zhang2023completionformer,wang2022depthhole}: an encoder–decoder network first fuses depth and RGB to predict a coarse dense depth, which is then refined by spatial propagation modules guided by RGB. Although effective on benchmarks, these models are highly domain- and pattern-specific, struggling under appearance shifts (lighting, texture, scene type) or heterogeneous real-world depth patterns such as sparsity, missing regions~\cite{wang2022depthhole}, and range limitations caused by sensor constraints.

In contrast to data-centric paradigms that train completion models from scratch on massive datasets~\cite{zuo2025omni,wang2025pacgdc,wang2023g2,ma2026metricanything}, recent efforts integrate MDE into depth completion to improve cross-domain robustness by combining the rich geometric priors of pretrained MDE with metric cues from sparse depth. 
Two families have emerged.
\textit{Test-time adaptation methods}~\cite{viola2025marigolddc,hyoseok2025zero,jeong2025testprompt, jeong2025test}, such as Marigold-DC~\cite{viola2025marigolddc} and TestPromptDC~\cite{jeong2025testprompt}, leverage sparse depth as supervision to iteratively optimize model outputs during inference, achieving high metric accuracy while preserving the MDE pipeline. 
However, such approaches incur heavy inference costs, limiting their real-time robotic applications.
In contrast, 
\textit{Direct feed-forward methods}~\cite{wang2025priorda,park2024depthprompt,yularge2026} prioritize efficiency but face the challenge of adapting relative depth models to metric prediction. 
PriorDA~\cite{wang2025priorda} adopts a two-stage pipeline: it first explicitly aligns the frozen MDE's relative output with sparse measurements to obtain a coarse dense map, then refines it using a secondary RGB-guided model. 
However, we argue that such \textit{output-level fusion} distorts MDE's geometric priors because the relative-to-metric mapping is often spatially inconsistent. This introduces structural artifacts that the refinement stage must laboriously correct, ultimately making the pipeline heavier and more sensitive to unseen depth patterns (detailed two-stage limitation analysis in Supp.~\ref{supsec:2stage}).
Meanwhile, PromptDA~\cite{lin2025promptingda}, designed for depth super-resolution, does not transfer well to depth completion due to its specific depth pattern.

To overcome these limitations, we rethink depth completion as a one-stage scale-prompting adaptation of MDE and design a scale-aware prompt encoder that constructs unified scale prompts for varying depth patterns, effectively integrating MDE’s geometric priors with sparse depth cues to achieve domain-general and pattern-agnostic performance.

\label{sec:method}
\section{Any2Full}
In this section, we present Any2Full, a one-stage depth completion framework that is domain-general and depth-pattern-agnostic. 
\cref{sec:method-formulation} introduces our scale-prompting formulation, which adapts a pretrained monocular depth estimation (MDE) model for depth completion by injecting scale cues from sparse inputs and leveraging its domain-general geometric priors.
\cref{sec:method-module} details the scale prompting pipeline, centered around a scale-aware prompt encoder that constructs unified scale prompts for diverse depth patterns.
Finally, \cref{sec:method-traning} describes training objectives and implementation details.


\subsection{Formulation}\label{sec:method-formulation}

Depth completion aims to recover a dense metric depth map $\mathbf{D}_f\in \mathbb{R}^{H \times W }$  from an RGB image $\mathbf{I} \in \mathbb{R}^{H \times W \times 3}$ and its corresponding sparse measurements $\mathbf{D}_s$.
Traditional approaches directly learn a mapping $\hat{\mathbf{D}}_f = \mathcal{F}_{\theta}(\mathbf{I}, \mathbf{D}_s)$ by fusing RGB and sparse depth, which often results in domain-specific representations.

\begin{figure*}[!t]
  \centering
  \includegraphics[width=0.9\textwidth]{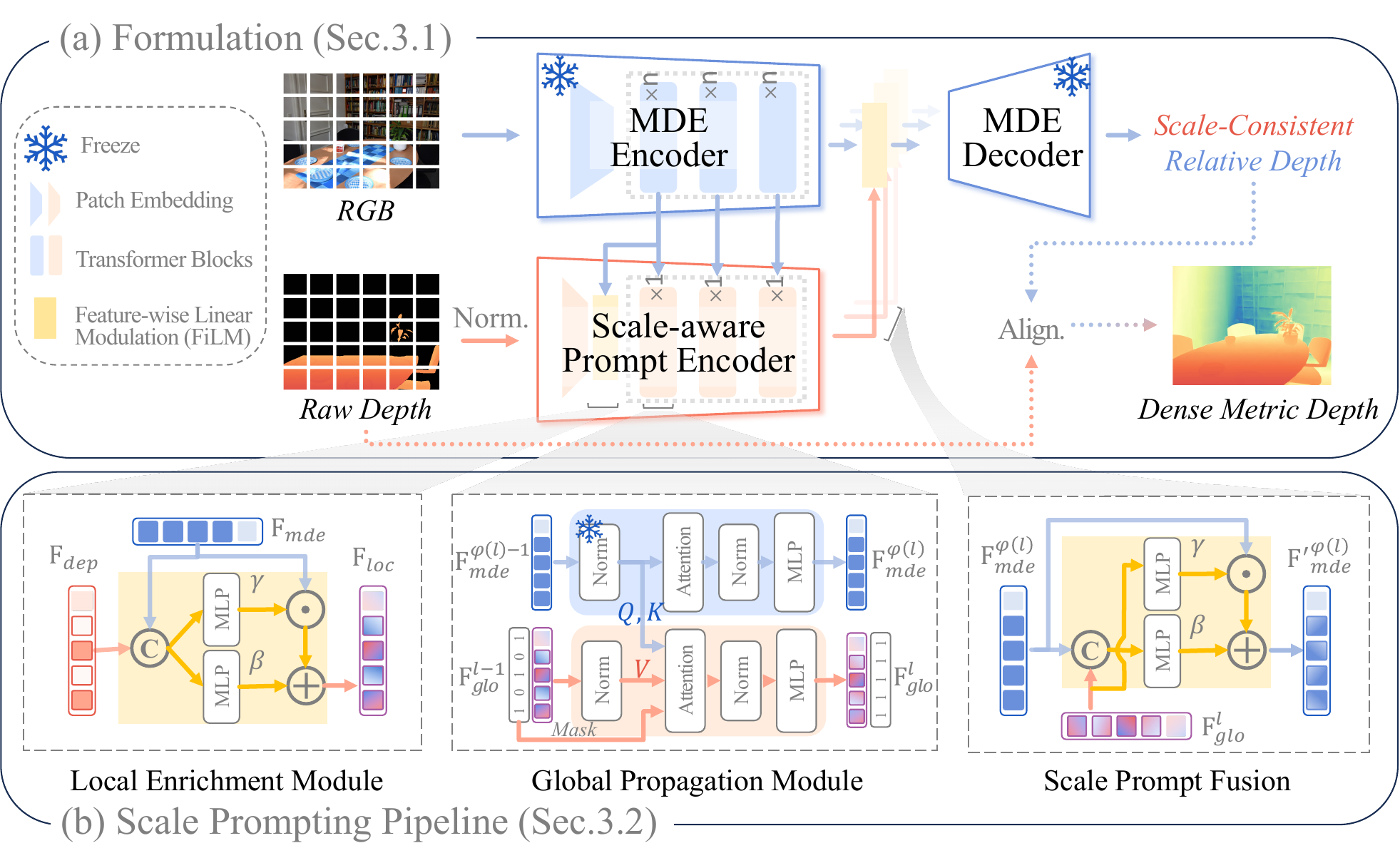}
  \caption{\textbf{Overview of Any2Full.} 
  (a) Our framework reformulates depth completion as a scale-prompting adaptation of a pretrained MDE model. The normalized raw depth is encoded into scale prompts to modulate MDE features for scale-consistent relative depth prediction, followed by a non-parametric least-squares fit to recover the dense metric depth. (b) The Scale-aware Prompt Encoder (SAPE) transforms raw depth into unified scale prompts through two hierarchical modules: \textit{Local Enrichment} anchors scale cues into the MDE latent space; and \textit{Global Propagation} leverages MDE geometric features as guidance to diffuse scale cues across patches. Finally, \textit{Scale Prompt Fusion} injects the resulting prompts into the MDE decoder for final prediction.}
  \label{fig:framework}
\end{figure*}

To overcome this limitation, we \textbf{reformulate} depth completion as a scale-prompting adaptation of pretrained MDE models. \cref{fig:framework} (a) illustrates our one-stage framework Any2Full, where depth inputs are encoded into scale prompts that modulate pretrained MDE features to produce \textit{scale-consistent} relative depth, which can be converted to metric depth using a single global scale and bias (as analyzed in Supp.~\ref{supsec:scale}).
This formulation enables the model to inherit the domain-general geometric priors of a pretrained MDE backbone.

Formally, given a pretrained MDE model $\mathcal{M}$, an RGB image $\mathbf{I}$ and a raw depth map $\mathbf{D}_s$, our goal is to estimate a dense metric depth $\hat{\mathbf{D}}_f$. To bridge the gap between sparse metric observations and the MDE's relative representation, we first normalize $\mathbf{D}_s$ into $\tilde{\mathbf{D}}_s$ by removing global scale and bias while preserving inter-point scale ratios as scale cues (details in Supp.~\ref{supsec:norm}). We then employ a \textbf{Scale-aware Prompt Encoder (SAPE)}, denoted as $\mathcal{G}$ to convert $\tilde{\mathbf{D}}_s$ into a scale prompt that modulates the features of $\mathcal{M}$ to produce a scale-consistent relative prediction  

\begin{equation}
\hat{\tilde{\mathbf{D}}}_f = \mathcal{M} \big( \mathbf{I} \mid \mathcal{G}(\tilde{\mathbf{D}}_s) \big).
\end{equation}

Finally, the relative prediction $\hat{\tilde{\mathbf{D}}}_f $ is aligned with the raw metric depth $\mathbf{D}_s$ via non-parametric least-squares fit~\cite{ranftl2020towards} to recover the dense metric depth $\hat{\mathbf{D}}_f$. 

\textit{One-stage inference.}
We define one-stage as a single forward inference process without intermediate depth prediction or auxiliary refinement networks.
Our framework satisfies this definition because the final metric depth is obtained via a closed-form alignment applied to the predicted relative depth, which introduces no additional learnable modules.

Overall, our formulation bridges depth completion and monocular depth estimation within a one stage scale-prompting framework preserving MDE's geometric priors while injecting scale cues for domain-general depth completion.

\subsection{Scale Prompting Pipeline}\label{sec:method-module}
The core challenge in designing the SAPE $\mathcal{G}$ within this scale prompting pipeline lies in handling varying depth patterns, which exhibit different sparsity levels and irregular spatial distributions (i.e., random missing regions), leading to unstable training and triggering pattern-specific overfitting.

To address this, SAPE progressively reconstructs the structural context of sparse depth input into a unified scale prompt representation under the
guidance of MDE’s geometric priors, with two hierarchical modules:
(1) a local enrichment module that produces local scale-aware features robust to sparsity variations, and
(2) a global propagation module that handles irregular spatial distributions and establishes a globally consistent representation as scale prompt.  
Then, a scale prompt fusion module uses the resulting scale prompt to modulate MDE features for scale-consistent prediction, forming our scale prompting pipeline. \cref{fig:framework} (b) details the components.

Throughout, we build upon a ViT-style~\cite{dosovitskiy2020image} MDE backbone~\cite{yang2024depthanythingv2} and the SAPE follows ViT encoder with a lightweight design. Given the sparse depth input $\tilde{\mathbf{D}}_s$, the SAPE first embeds it into patch-level depth features $\mathbf{F}_{dep}$ (see Supp.~\ref{supsec:embed} for details). These features are subsequently encoded into unified scale prompts via the following two hierarchical modules:

\vspace{4pt}
\noindent\textbf{Local Enrichment.}
To achieve robustness under varying sparsity levels, the Local Enrichment module distills scale cues from $\mathbf{F}_{dep}$ and couples them with the dense geometric features $\mathbf{F}_{mde}$ from the MDE backbone, producing local scale-aware features $\mathbf{F}_{loc}$ that are insensitive to sparsity variations.

In detail, each patch-level depth feature $\mathbf{f}_{dep,i}$ interacts with its corresponding MDE feature $\mathbf{f}_{mde,i}$ through generalized Feature-wise Linear Modulation (FiLM)~\cite{perez2018film} mechanism to anchor scale cues within the MDE latent space:

\begin{equation}
\mathbf{f}_{loc,i} = \gamma(\mathbf{f}_{dep,i}, \mathbf{f}_{mde,i}) \odot \mathbf{f}_{mde,i} + \beta(\mathbf{f}_{dep,i}, \mathbf{f}_{mde,i}),
\end{equation}
\noindent where $\odot$ denotes element-wise multiplication, $\gamma(\cdot)$ and $\beta(\cdot)$ are modulation parameters predicted by a lightweight Multi-Layer Perceptron (MLP).

\vspace{4pt}
\noindent\textbf{Global Propagation.}
While the Local Enrichment module anchors scale cues within individual patches, these cues remain spatially fragmented and lack global consistency. To bridge this gap, the Global Propagation module performs MDE geometry-guided diffusion.

Specifically, we treat the local scale-aware features $\mathbf{F}_{loc}$ as the initial state $\mathbf{F}_{glo}^{0} = \mathbf{F}_{loc}$, and update them through $L$ geometry-guided Transformer blocks, where $L$ matches the MDE encoder groups (and corresponding decoder levels):
\begin{equation}
\mathbf{F}_{glo}^{l} = \text{TransformerBlock}(\mathbf{F}_{glo}^{l-1}, \mathbf{F}_{mde}^{\phi(l)-1}), \;l=1\!:\!L.
\end{equation}
where $\mathbf{F}_{mde}^{\phi(l)-1}$ denotes the geometry features extracted from the last block of the $l$-th MDE encoder group, and $\phi(l) = l \cdot n $ is the index mapping function (assuming each group contains $n$ blocks).
Different from standard cross-attention mechanisms, our module computes attention weights purely from MDE geometry (Query and Key from $\mathbf{F}_{mde}$), using $\mathbf{F}_{glo}$ only as the Value.  This ensures that scale cues are diffused along geometric structures defined by the RGB-based MDE features rather than being biased by the irregular sampling patterns of the sparse depth. 
The resulting globally consistent scale-aware features $\mathbf{F}_{glo}^{l}$ provide a unified scale prompt for diverse depth patterns that benefit subsequent prompt-based fusion in a pattern-agnostic manner. 

\vspace{4pt}
\noindent\textbf{Scale Prompt Fusion.}
The set of scale prompts $\{\mathbf{F}_{glo}^{1}, \dots, \mathbf{F}_{glo}^{L}\}$ is injected into the MDE decoder for scale-consistent prediction through hierarchical Feature-wise Linear Modulation (FiLM). 
Specifically, at each decoder level $l$, the modulation is applied to $\mathbf{F}_{mde}^{\phi(l)}$, which is the intermediate MDE feature within the decoder path prior to its fusion with other level features:
\begin{equation}
\mathbf{F'}_{mde}^{\phi(l)}
= \gamma^{l} \!\left( \mathbf{F}_{glo}^{l} \right) \mathbf{F}_{mde}^{\phi(l)} 
+ \beta^{l} \!\left( \mathbf{F}_{glo}^{l}, \mathbf{F}_{mde}^{\phi(l)} \right),
\; l=1\!:\!L.
\end{equation}
$\gamma^{l}(\cdot)$ and $\beta^{l}(\cdot)$ are predicted by lightweight MLP specialized for each level. 
Each modulation is applied token-wise within its level.   
This multi-level fusion injects scale prompt at multiple semantic levels, refining relative depth features toward scale-consistent predictions while preserving the domain-general geometric priors from the MDE backbone.

\vspace{4pt}
\noindent\textbf{Efficiency and Lightweight Design.} The Scale-aware Prompt Encoder is designed for minimal overhead while ensuring effective guidance.
To achieve this, the number of propagation blocks is specifically aligned with the number of hierarchical levels in the MDE decoder, ensuring that each scale prompt semantically corresponds to a specific decoder level without redundant computation.
Moreover, propagation and prompting are omitted for the lowest-level encoder-decoder layers, where features primarily capture local textures rather than the structural geometric priors necessary for scale propagation.\footnote{\scriptsize For example, with a ViT-L backbone~\cite{yang2024depthanythingv2} providing intermediate features at layers $\{5, 11, 17, 23\}$, we enrich local features using layer $10$, perform three propagation steps guided by layers $\{10, 16, 22\}$, and execute prompt fusion at levels corresponding to layers $\{11, 17, 23\}$.}
To prevent early-propagation loss of sparse scale cues, we adopt masked attention in the first transformer block, constraining scale information spread from valid depth locations to geometrically coherent regions.

\vspace{4pt}
\noindent\textbf{Design Analysis.}
\textit{Pattern-agnostic. }Scale-aware Prompt Encoder adapts to diverse depth patterns. The depth input provides only scale cues at valid points, while enrichment and propagation rely on MDE geometry, ensuring invariance to sparsity levels and spatial distributions.
\textit{Domain-general.} Unlike conventional prompting methods~\cite{jia2022visual,li2021prefix} that inject task-specific tokens at the input level to condition model behavior, our approach applies lightweight feature-level modulation via FiLM using a scale prompt. This design minimally alters the original MDE representations, injecting only scale-related cues to enhance scale consistency while preserving domain-general priors.


\subsection{Training}\label{sec:method-traning}
\textbf{Training Data.}
Since Any2Full requires RGB-D pairs with accurate metric ground truth, we train our model on high-quality synthetic datasets. To encourage robustness to varying depth patterns, we adopt two depth sampling strategies: (1) Random Sampling, which randomly selects depth points to produce sparse depth maps with varying density; and (2) Hole Sampling, following the method of ~\cite{wang2022depthhole}, produces depth inputs with large contiguous missing regions. During training, we randomly select one of the two strategies for each sample to ensure pattern diversity.

\noindent\textbf{Training Objective.}
We supervise the predicted dense depth using a combination of losses that enforce global scale consistency, local structure sharpness, and sparse input alignment.
Following prior MDE works~\cite{ranftl2020towards,yang2024depthanythingv1,wang2024depthholev2}, we adopt a scale- and shift-invariant loss $\mathcal{L}_{\text{ssi}}$ for global consistency and a gradient matching loss $\mathcal{L}_{\text{gm}}$ for edge preservation.
To ensure consistency with sparse measurements, we additionally apply $\mathcal{L}_{\text{ssi}}$ on valid depth anchors, denoted as $\mathcal{L}_{\text{anchor}}$.
We also include a Relative-Structure SSIM loss $\mathcal{L}_{r\text{-ssim}}$~\cite{hyoseok2025zero} to regularize structural similarity between prediction and ground truth.
The overall training objective is a weighted sum of these losses (see Supp.~\ref{supsec:loss} for details).

\noindent\textbf{Implementation Details}. We utilize \textit{Depth Anything v2}~\cite{yang2024depthanythingv2} as the MDE backbone, initialized from its released relative-depth model. We freeze the MDE backbone and train the scale-aware prompt encoder on a subset of the Depth Anything v2 synthetic dataset, including 60K indoor images from \textit{Hypersim}~\cite{roberts2021hypersim}, 10K outdoor images sampled from \textit{VKITTI2}~\cite{cabon2020vkitti}, and 15K images sampled from \textit{TartanAir}~\cite{wang2020tartanair} covering both indoor and outdoor scenes. The sampling is conducted to maintain dataset balance and diversity while considering computational constraints. The model is trained for 224K steps with a 10K-step warm-up phase and a cosine scheduler~\cite{loshchilov2016sgdr}. We use a batch size of 16 on four NPUs (equivalent to high-end GPU clusters). The optimizer is Adam~\cite{kingma2014adam} with a learning rate of 5e\text{-}5 for the scale-aware prompt encoder.

\section{Experiments}
\label{sec:experiment}
\subsection{Experimental Setup}
\textbf{Public Benchmarks.}
To assess the zero-shot generalization ability of Any2Full across visual domains, we evaluate on six unseen public datasets spanning indoor and outdoor scenes with diverse visual conditions and sensor modalities, forming a comprehensive benchmark for cross-domain depth completion.
\textbf{NYU-Depth V2}~\cite{silberman2012nyuv2} contains indoor scenes captured by an RGB-D Kinect sensor, and we follow the standard test split of 654 samples at 304×228 resolution.
\textbf{iBims-1}~\cite{koch2018ibims} is a high-quality indoor dataset captured with a laser scanner featuring accurate geometry and an extended depth range of up to 50 m, and we use all 100 test images at 640×480 resolution. 
\textbf{KITTI DC}~\cite{uhrig2017kittidc} consists of outdoor driving scenes with sparse LiDAR depth at 1216×352 resolution, and we adopt the standard validation split of 1,000 samples.
\textbf{DIODE}~\cite{vasiljevic2019diode} is a large-scale indoor–outdoor dataset captured with a FARO laser scanner, providing highly accurate dense depth up to 350 m. We use the official validation split of 771 samples at 640×480 resolution.
\textbf{ETH3D}~\cite{schops2017eth3d} is a multi-view stereo benchmark featuring high-resolution indoor and outdoor scenes from DSLR imagery and FARO scans, and we evaluate the High-res DSLR set (11 scenes, 390 images) at 640×480 resolution.
\textbf{VOID}~\cite{wong2020unsupervised} contains indoor sequences captured by an Intel RealSense D435i sensor, featuring sparse, noisy depth with challenges such as motion blur and low texture. We use the standard test set of 800 samples with three sparsity protocols (1500/500/150) at 640×480 resolution.

\noindent\textbf{Real-World Dataset.}
Beyond public benchmarks, we further evaluate on a newly collected real-world dataset \textbf{Logistic-Black} to assess the practical robustness of our model. 
This dataset is collected in a robotic warehouse scene using a Vzense DS77C Pro ToF camera. It contains 114 RGB-D pairs cropped to 1448×1048 resolution. Logistic-Black represents a challenging robotic–industrial domain characterized by black packages with top-down camera views. Black packages absorb infrared light, leading to missing regions in ToF depth measurements. 
To obtain metrically accurate ground truth for evaluation, we replace the package surfaces with reflective materials under the same setup. 


\noindent\textbf{Depth Patterns.}
To comprehensively evaluate the robustness of Any2Full across different depth patterns, we design six representative settings: Hole, Range, Sparse-Random, Sparse-LiDAR, Sparse-SfM, and Mixed. Sparse patterns follow each dataset’s standard protocol. Hole and Range patterns follow and extend robustness protocols in prior studies~\cite{wang2022depthhole, wang2025priorda}. Range pattern is unseen for all methods, as none are trained with range-truncated depth inputs.
\textbf{Hole.} Simulates missing regions commonly observed in indoor RGB-D sensors, where raw depth is dense, but continuous areas are missing due to reflection or absorption of sensor light. Following~\cite{wang2022depthhole,wang2024depthholev2}, we adopt highlight masking, black masking, small-block masking, and semantic masking to generate realistic hole distributions.
\textbf{Range.} Models sensor range limitations~\cite{park2024depthprompt,wang2025priorda}. TOF sensors in indoor scenes are typically unreliable at very near or far distances, while outdoor LiDARs are range-limited on reflective surfaces. We retain the central 20–80\% of valid depth values for both indoor and outdoor, forming a range-constrained completion task.
\textbf{Sparse-Random.} This widely used setting randomly samples sparse points from dense ground truth, though it differs from real-world depth distributions. Following~\cite{ma2018sparse,cheng2018depthcspn,park2020nonlocalspn}, we randomly sample 500 points for NYU-Depth V2, 1 000 points for iBims-1 and 0.1\% points for DIODE.
\textbf{Sparse-LiDAR.} This pattern corresponds to outdoor depth captured by LiDAR sensors, we use the original 64-lines LiDAR measurements (64L) as sparse inputs for KITTI DC. 
\textbf{Sparse-SfM.} Following~\cite{zuo2025omni}, we project COLMAP SfM points into image space to obtain sparse depth maps representative of multi-view reconstruction for ETH3D.
\textbf{Mixed.} This pattern is specific to the \textit{Logistic-Black} dataset, combining depth holes caused by low-reflectivity black packages and sparsity induced by the ToF sensor's limited depth resolution ($\sim$ 20\%).

\noindent\textbf{Baselines.} 
We compare Any2Full with three method categories:
(1) Domain-specific depth completion: CompletionFormer (CompFormer)~\cite{zhang2023completionformer} and DepthPrompting (DepthPrompt)~\cite{park2024depthprompt}, where the abbreviations are used for compactness.
(2) Domain-general monocular depth estimation (MDE): Depth Anything v2~\cite{yang2024depthanythingv2} and Marigold~\cite{ke2024marigold}, which predict relative depth from RGB only.
(3) Domain-general methods: PromptDA~\cite{lin2025promptingda} is designed for depth super-resolution. Marigold-DC~\cite{viola2025marigolddc}, TestPromptDC~\cite{jeong2025testprompt}, PriorDA~\cite{wang2025priorda}, and OMNI-DC~\cite{zuo2025omni} are depth completion methods, where all except OMNI-DC leverage MDE priors.

\noindent\textbf{Evaluation Protocols.}
All models are evaluated in a zero-shot setting without dataset-specific fine-tuning. 
For domain-specific methods, we use their released models trained on the NYU-Depth V2 dataset for evaluation.
For MDE methods, relative depth predictions are aligned to sparse depth using least-squares fitting to produce metric outputs.
All Depth Anything (DA)–based methods use the released DA-v2-Large backbone for consistency. For PromptDA, the raw sparse depth is downsampled via nearest-neighbor interpolation to form dense low-resolution inputs. We compute Absolute Relative Error (AbsREL) and Root Mean Squared Error (RMSE, in meters)~\cite{koch2018evaluation} on valid pixels under the same cross-domain and depth-pattern settings.

\begin{table*}[t]
\centering
\caption{%
 \textbf{Quantitative comparison on large-scale zero-shot benchmarks} 
(Metrics are AbsREL (↓) and RMSE (↓); 
\best{bold} and \second{underlined} denote the best and second-best results).
 “Any2Full (DA-L/B/S)” denotes our model
using Depth Anything-Large/Base/Small backbones. 
\textbf{Our method Any2Full (DA-L) achieves the lowest average rank (2.3) across all scenarios},
consistently performing among the top competitors against strong
domain-generalization and pattern-agnostic approaches,
demonstrating robust generalization across diverse depth patterns and domains.
}
\label{tab:zero-dc-quan}
\vspace{-6pt}
\begin{threeparttable}
\newcolumntype{M}{>{\centering\arraybackslash}p{1.1cm}}
\setlength{\tabcolsep}{2pt}

\resizebox{\textwidth}{!}{%
\begin{tabular}{l >{\bfseries\centering\arraybackslash}p{1.5cm} *{6}{M} !{\vrule width 0.8pt} *{6}{M} *{6}{M}}
\toprule
 & \textbf{Avg} & \multicolumn{6}{c}{\textbf{ AVG }} & \multicolumn{6}{c}{\textbf{NYU-Depth V2}} & \multicolumn{6}{c}{\textbf{IBims-1}} \\
\cmidrule(lr){3-8}\cmidrule(lr){9-14}\cmidrule(lr){15-20}
 & \textbf{Rank} & \multicolumn{2}{c}{Hole} & \multicolumn{2}{c}{Range} & \multicolumn{2}{c}{Sparse} & \multicolumn{2}{c}{Hole} & \multicolumn{2}{c}{Range} & \multicolumn{2}{c}{Sparse-Random} & \multicolumn{2}{c}{Hole} & \multicolumn{2}{c}{Range} & \multicolumn{2}{c}{Sparse-Random} \\
\cmidrule(lr){3-4}\cmidrule(lr){5-6}\cmidrule(lr){7-8} \cmidrule(lr){9-10}\cmidrule(lr){11-12}\cmidrule(lr){13-14} \cmidrule(lr){15-16}\cmidrule(lr){17-18}\cmidrule(lr){19-20}
\multirow{-3}{*}{Method} & $\downarrow$ & AbsREL & RMSE & AbsREL & RMSE & AbsREL & RMSE & AbsREL & RMSE & AbsREL & RMSE & AbsREL & RMSE & AbsREL & RMSE & AbsREL & RMSE & AbsREL & RMSE \\
\midrule
CompFormer & 
7.9 & 0.316 & 9.682 & 0.355 & 5.499 & 0.067 & 8.020 & 
0.342 & 1.098 & 0.324 & 1.050 & \best{0.011} & \best{0.089} & 
0.371 & 1.372 & 0.372 & 1.511 & 0.013 & 0.197 \\
DepthPrompt$^{\ddagger \bigstar}$ & 
5.1 & 0.018 & 0.816 & 0.121 & 3.558 & 0.098 & 2.032 & 
\second{0.008} & \third{0.097} & \second{0.051} & \second{0.338} & \third{0.014} & \second{0.104} & 
\second{0.008} & 0.145 & 0.091 & 0.647 & 0.017 & 0.202 \\
OMNI-DC$^{\dagger \ddagger}$ & 
\cellcolor{silver}{3.4} & \second{0.014} & \third{0.472} & 0.082 & 2.805 & \best{0.025} & \best{0.648} & 
0.019 & 0.108 & 0.073 & 0.415 & \second{0.014} & \third{0.112} & 
0.014 & 0.152 & 0.076 & 0.592 & \best{0.009} & \second{0.148} \\

Depth Anything$^{\dagger \bigstar}$ & 
6.4 & 0.069 & 2.361 & 0.079 & 2.472 & 0.074 & 2.207 & 
0.062 & 0.313 & 0.074 & \third{0.342} & 0.061 & 0.299 & 
0.038 & 0.329 & \third{0.042} & \best{0.303} & 0.038 & 0.337 \\
Marigold$^{\dagger \bigstar}$ & 
7.5 & 0.115 & 2.084 & 0.294 & 2.665 & 0.142 & 2.230 & 
0.080 & 0.324 & 0.099 & 0.375 & 0.082 & 0.320 & 
0.060 & 0.346 & 0.075 & 0.400 & 0.060 & 0.354 \\
PromptDA$^{\dagger \bigstar}$& 
9.1 & 0.400 & 6.036 & 0.644 & 8.063 & 0.408 & 5.643 & 
0.333 & 1.349 & 0.588 & 2.240 & 0.033 & 0.224 & 
0.238 & 1.232 & 0.533 & 2.634 & 0.021 & 0.191 \\
Marigold-DC~$^{\dagger \ddagger \bigstar}$& 
5.7 & 0.024 & 0.597 & 0.091 & 2.995 & 0.193 & 2.088 & 
0.017 & 0.114 & 0.080 & 0.412 & 0.021 & 0.153 & 
0.016 & 0.155 & 0.070 & 0.535 & 0.018 & 0.208 \\
TestPromptDC$^{\dagger \ddagger \bigstar}$ & 
\cellcolor{bronze}{3.5} & 0.017 & \best{0.445} & \second{0.049} & \best{1.660} & 0.161 & 2.046 & 
0.014 & \second{0.079} & \third{0.057} & 0.393 & 0.018 & 0.114 & 
0.013 & \second{0.120} & \second{0.038} & \second{0.306} & 0.015 & 0.163 \\
PriorDA$^{\dagger \ddagger \bigstar}$& 
3.6 & 0.015 & 0.655 & 0.080 & 2.567 & 0.028 & \second{0.775} & 
\third{0.013} & 0.102 & 0.074 & 0.363 & 0.016 & 0.119 & 
\third{0.011} & \third{0.141} & 0.063 & 0.459 & \second{0.011} & \best{0.143} \\
\midrule
\rowcolor{oursrow}
Any2Full(DA-L)$^{\dagger \ddagger \bigstar}$ & 
\textbf{\cellcolor{gold}{2.3}} & \best{0.010} & \second{0.459} & \best{0.046} & \second{2.184} & \second{0.026} &\third{0.829} & 
\best{0.008} & \best{0.067} & \best{0.035} & \best{0.259} & 0.017 & 0.138 & 
\best{0.007} & \best{0.110} & \best{0.030} & \third{0.334} & \third{0.013} & \third{0.161} \\

\rowcolor{oursrow}
\leavevmode\color{gray} Any2Full(DA-B)$^{\dagger \ddagger \bigstar}$ &  & \leavevmode\color{gray} 0.010 & \leavevmode\color{gray}0.445 &  \leavevmode\color{gray} 0.046 &  \leavevmode\color{gray} 2.134 &  \leavevmode\color{gray} 0.028 &  \leavevmode\color{gray} 0.822 \leavevmode\color{gray}  & 
 \leavevmode\color{gray} 0.008 &  \leavevmode\color{gray} 0.066 &  \leavevmode\color{gray} 0.035 &  \leavevmode\color{gray} 0.264 &  \leavevmode\color{gray} 0.017 &  \leavevmode\color{gray} 0.131 & 
 \leavevmode\color{gray} 0.008 &  \leavevmode\color{gray} 0.112 &  \leavevmode\color{gray} 0.031 &  \leavevmode\color{gray} 0.336 &  \leavevmode\color{gray} 0.013 &  \leavevmode\color{gray} 0.167 \\

\rowcolor{oursrow}
\leavevmode\color{gray} Any2Full(DA-S)$^{\dagger \ddagger \bigstar}$ &  & \leavevmode\color{gray} 0.011 & \leavevmode\color{gray} 0.484 & \leavevmode\color{gray} 0.0549 & \leavevmode\color{gray} 2.531 & \leavevmode\color{gray} 0.029 & \leavevmode\color{gray} 0.877 & \leavevmode\color{gray} 0.008 & \leavevmode\color{gray} 0.068 & \leavevmode\color{gray} 0.041 & \leavevmode\color{gray} 0.303 & \leavevmode\color{gray} 0.017 & \leavevmode\color{gray} 0.126 & \leavevmode\color{gray} 0.009 & \leavevmode\color{gray} 0.126 & \leavevmode\color{gray} 0.039 & \leavevmode\color{gray} 0.405 & \leavevmode\color{gray} 0.015 & \leavevmode\color{gray} 0.174 \\
\bottomrule
\end{tabular}}

\vspace{0.0cm}

\resizebox{\textwidth}{!}{%
\begin{tabular}{l *{6}{M} *{6}{M} *{6}{M} *{2}{M}}
\toprule
 & \multicolumn{6}{c}{\textbf{DIODE}} & \multicolumn{6}{c}{\textbf{KITTI DC}} & \multicolumn{6}{c}{\textbf{ETH3D}} & \multicolumn{2}{c}{\textbf{Logistic-Black}} \\
\cmidrule(lr){2-7}\cmidrule(lr){8-13}\cmidrule(lr){14-19}\cmidrule(lr){20-21}
 & \multicolumn{2}{c}{Hole} & \multicolumn{2}{c}{Range} & \multicolumn{2}{c}{Sparse-Random} & \multicolumn{2}{c}{Hole} & \multicolumn{2}{c}{Range} & \multicolumn{2}{c}{Sparse-LiDAR} & \multicolumn{2}{c}{Hole} & \multicolumn{2}{c}{Range} & \multicolumn{2}{c}{Sparse-SfM} & \multicolumn{2}{c}{Mixed} \\
\cmidrule(lr){2-3}\cmidrule(lr){4-5}\cmidrule(lr){6-7} \cmidrule(lr){8-9}\cmidrule(lr){10-11}\cmidrule(lr){12-13} \cmidrule(lr){14-15}\cmidrule(lr){16-17}\cmidrule(lr){18-19} \cmidrule(lr){20-21}
Method & AbsREL & RMSE & AbsREL & RMSE & AbsREL & RMSE & AbsREL & RMSE & AbsREL & RMSE & AbsREL & RMSE & AbsREL & RMSE & AbsREL & RMSE & AbsREL & RMSE & AbsREL & RMSE \\
\midrule
CompFormer & 
0.353 & 15.110 & 0.384 & 7.617 & 0.034 & 11.014 & 
0.163 & 28.108 & 0.307 & 13.370 & 0.041 & 18.348 & 
0.350 & 2.724 & 0.387 & 3.946 & 0.235 & 10.450 & 0.011 & 0.039 \\
DepthPrompt$^{\ddagger \bigstar}$ & 
\third{0.020} & 1.091 & 0.152 & 4.494 & 0.029 & 1.315 & 
0.040 & 2.428 & 0.170 & 9.967 & 0.098 & 5.005 & 
0.012 & 0.319 & 0.142 & 2.343 & 0.332 & 3.534 & \second{0.009} & \second{0.025} \\
OMNI-DC$^{\dagger \ddagger}$ & 
\best{0.015} & \best{0.962} & 0.100 & 3.839 & \best{0.017} & \best{0.915} & 
\second{0.011} & \third{0.907} & \third{0.079} & 7.469 & \best{0.013} & \best{1.230} & 
0.010 & 0.231 & 0.081 & 1.710 & \third{0.070} & \second{0.835} & \best{0.009} & \best{0.022} \\

Depth Anything$^{\dagger \bigstar}$ & 
0.110 & 5.061 & 0.130 & 5.083 & 0.110 & 4.937 & 
0.078 & 3.507 & 0.090 & \second{4.646} & 0.085 & 
3.830 & 0.058 & 2.594 & \third{0.062} & 1.988 & 0.075 & 1.633 & 0.026 & 0.059 \\
Marigold$^{\dagger  \bigstar}$ & 
0.140 & 2.751 & 0.140 & \third{3.446} & 0.141 & 2.770 & 
0.203 & 6.080 & 0.206 & 7.951 & 0.194 & 6.192 & 
0.092 & 0.918 & 0.950 & \third{1.152} & 0.232 & 1.513 & 0.025 & 0.061 \\
PromptDA$^{\dagger \bigstar}$& 
0.299 & 7.244 & 0.567 & 10.831 & 0.038 & 1.691 & 
0.829 & 16.804 & 0.916 & 18.710 & 0.967 & 18.960 & 
0.301 & 3.549 & 0.614 & 5.898 & 0.979 & 7.149 & 0.990 & 1.714 \\
Marigold-DC$^{\dagger \ddagger \bigstar}$ & 
0.034 & \second{1.028} & 0.099 & 3.755 & 0.034 & 
1.260 & 0.041 & 1.489 & 0.105 & 8.516 & 0.044 & 
1.865 & 0.013 & 0.198 & 0.100 & 1.755 & 0.847 & 6.954 & 0.011 & 0.036 \\

TestPromptDC$^{\dagger \ddagger \bigstar}$ & 
0.034 & \third{1.073} & \second{0.085} & \second{3.315} & 0.033 & \second{1.138} & 
\third{0.016} & \second{0.836} & \best{0.037} & \best{3.630} & \third{0.020} & \second{1.350} & 
\third{0.008} & \best{0.118} & \best{0.032} & \best{0.654} & 0.719 & 7.465 & 0.013 & 0.040 \\
PriorDA$^{\dagger \ddagger \bigstar}$ & 
0.024 & 1.218 & \third{0.091} & 3.586 & \second{0.022} & \third{1.144} & 
0.020 & 1.623 & 0.096 & 6.990 & 0.022 & 1.714 & 
\second{0.007} & \third{0.190} & 0.078 & 1.438 & \second{0.070} & \best{0.754} & 0.011 & 0.037 \\
\midrule

\rowcolor{oursrow}
Any2Full(DA-L)$^{\dagger \ddagger \bigstar}$ & 
\second{0.018} & 1.164 & \best{0.062} & \best{3.217} & \third{0.022} & 1.363 & 
\best{0.011} & \best{0.784} & \second{0.061} & \third{6.000} & \second{0.017} & \third{1.356} & 
\best{0.005} & \second{0.173} & \second{0.041} & \second{1.111} & \best{0.063} & \third{1.129} & \third{0.010} & \third{0.029} \\

\rowcolor{oursrow}
\leavevmode\color{gray} Any2Full(DA-B)$^{\dagger \ddagger \bigstar}$ & \leavevmode\color{gray} 0.018 & \leavevmode\color{gray} 1.116 & \leavevmode\color{gray} 0.063 & \leavevmode\color{gray} 3.297 & \leavevmode\color{gray} 0.022 & \leavevmode\color{gray} 1.312 & \leavevmode\color{gray} 0.011 & \leavevmode\color{gray} 0.771 & \leavevmode\color{gray} 0.058 & \leavevmode\color{gray} 5.597 & \leavevmode\color{gray} 0.017 & \leavevmode\color{gray} 1.363 & \leavevmode\color{gray} 0.005 & \leavevmode\color{gray} 0.161 & \leavevmode\color{gray} 0.041 & \leavevmode\color{gray} 1.175 & \leavevmode\color{gray} 0.069 & \leavevmode\color{gray} 1.135 & \leavevmode\color{gray} 0.009 & \leavevmode\color{gray} 0.028 \\

\rowcolor{oursrow}
\leavevmode\color{gray} Any2Full(DA-S)$^{\dagger \ddagger \bigstar}$ & \leavevmode\color{gray} 0.019 & \leavevmode\color{gray} 1.257 & \leavevmode\color{gray} 0.068 & \leavevmode\color{gray} 3.422 & \leavevmode\color{gray} 0.023 & \leavevmode\color{gray} 1.429 & \leavevmode\color{gray} 0.012 & \leavevmode\color{gray} 0.794 & \leavevmode\color{gray} 0.070 & \leavevmode\color{gray} 7.043 & \leavevmode\color{gray} 0.018 & \leavevmode\color{gray} 1.450 & \leavevmode\color{gray} 0.006 & \leavevmode\color{gray} 0.178 & \leavevmode\color{gray} 0.053 & \leavevmode\color{gray} 1.483 & \leavevmode\color{gray} 0.070 & \leavevmode\color{gray} 1.205 & \leavevmode\color{gray} 0.010 & \leavevmode\color{gray} 0.030 \\
\bottomrule
\end{tabular}}

\begin{tablenotes}
    \tiny 
    \item $^{\dagger}$ Domain Generalization method.
    $^{\ddagger}$ Depth pattern-agnostic method.
    $^{\bigstar}$ MDE-based approach.
\end{tablenotes}

\end{threeparttable} 
\vspace{-5pt}
\end{table*}

\vspace{-5pt}
\subsection{Comparisons with the State of the Art}
\vspace{-2pt}
\textbf{Domain Generalization.} 
As shown in \cref{tab:zero-dc-quan},
the performance of domain-specific methods degrades sharply when transferred to unseen domains, as CompFormer (NYU→KITTI), revealing that traditional RGBD-fused approaches are overfit to a specific domain. 
Monocular depth estimation methods like DA and Marigold generalize well for relative depth prediction, but their scale inconsistency causes significant errors when aligning with metric inputs.

OMNI-DC (rank 3.4) improves domain generalization by leveraging large-scale multi-domain training on diverse synthetic datasets combined with scale normalization. 
Meanwhile, MDE-integrated depth completion methods, including Marigold-DC (rank 5.7), TestPromptDC (rank 3.5), and PriorDA (rank 3.6), enhance cross-domain robustness by incorporating geometric priors from pretrained MDE models.

In comparison, our one-stage MDE-integrated framework Any2Full further improves both stability and accuracy, achieving the lowest average rank (2.3) across all evaluated scenarios, consistently remaining among the top performers on every dataset. Rather than relying on large-scale training data, our scale-prompting strategy effectively fuses geometric priors from pretrained MDE with scale cues from sparse depth, leading to robust domain-general depth completion.

\begin{figure*}[t]
  \centering
  \includegraphics[width=1\textwidth]{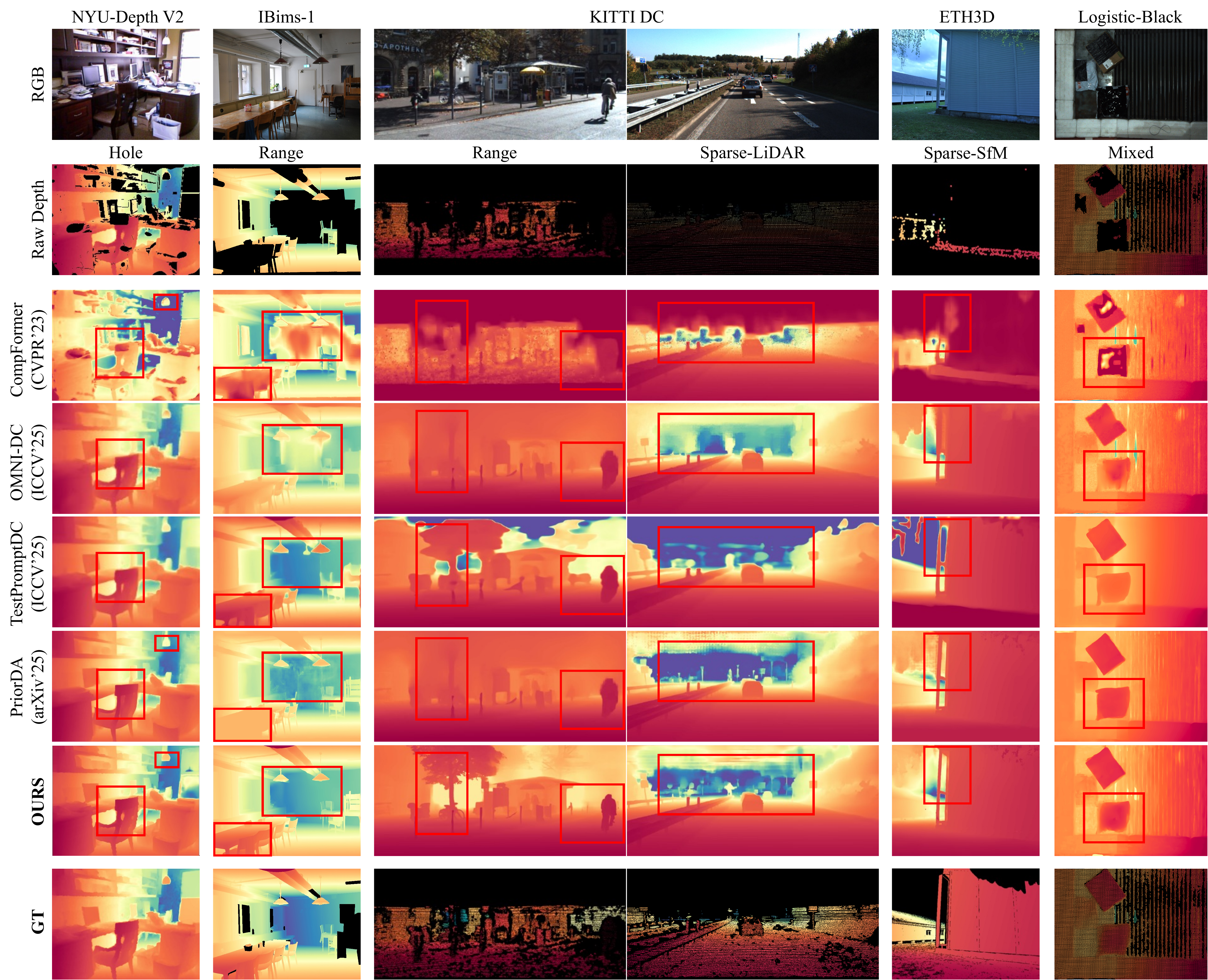}
  \vspace{-5pt}
  \caption{\textbf{Qualitative comparison} under various depth sampling patterns. Black shows missing depths, and red boxes mark key areas. \textbf{Any2Full demonstrates accurate global geometry, structural consistency, and fine-grained detail preservation across all patterns.} }
  \label{fig:zero-dc-qual}
\vspace{-10pt}
\end{figure*}

\vspace{4pt}
\noindent\textbf{Depth Pattern Generalization.}
As shown by the average metrics across various patterns in \cref{tab:zero-dc-quan}, existing methods exhibit high sensitivity to depth sampling patterns compared to Any2Full.
Among state-of-the-art pattern-agnostic methods, OMNI-DC and PriorDA appear to overfit to the Sparse setting and generalize poorly to unseen Range patterns, indicating their completion relies heavily on pattern-specific correlations. Meanwhile, test-time adaptation methods like TestPromptDC and Marigold-DC prove unstable in sparse scenarios, suffering significant performance degradation. \textbf{Qualitative Results} in \cref{fig:zero-dc-qual} further illustrate these differences. 
Traditional methods like CompFormer fail entirely. OMNI-DC exhibits noticeable depth bleeding (e.g., in Range and Sparse-SfM). While MDE-based methods benefit from fine-grained details, they introduce distinct artifacts: TestPromptDC produces erroneous depth rings and over-smoothing  (i.e., KITTI DC-Range) via its test-time prompt optimization and PriorDA generates artifacts (i.e., IBims-Range) from noisy output-level alignment with MDE predictions. In contrast, Any2Full maintains the fine-grained geometric details provided by the MDE while achieving superior global scale accuracy and structural consistency across all patterns. Furthermore, in the real-world dataset Logistic-Black, Any2Full achieves the most robust visual quality despite OMNI-DC's slightly lower AbsREL, benefiting downstream robotic tasks. These results confirm that our scale-aware prompt encoder effectively decouples scale cues from sparse depth, enabling pattern-invariant scale prompting.

\vspace{4pt}
\noindent\textbf{Model Efficiency.} As summarized in \cref{tab:depth_comparison}, Any2Full achieves state-of-the-art performance while maintaining high computational efficiency.
When built on DA-L, our method adds less than 20\% additional parameters (60.6M vs. 335.3M) to the MDE backbone, yet reduces AbsREL from 0.038 (Depth Anything) to 0.007. Compared to the competitive PriorDA using the same backbone, Any2Full achieves a \textbf{1.4$\times$} speedup (0.49 s vs. 0.68 s) with superior accuracy. 
Remarkably, the smallest variant of Ours (DA-S) still surpasses all prior depth completion methods, achieving a 0.09 s inference time, about \textbf{7$\times$} faster than PriorDA and \textbf{1000$\times$} faster than TestPromptDC. This highlights the efficiency of our one-stage scale prompting framework, which avoids the multi-stage refinement overhead of PriorDA or the iterative optimization required by TestPromptDC.

\begin{table}[t]
\centering
\caption{\textbf{Model efficiency comparison} on IBims-1.
Average runtime is computed on 100 640×480 images and hole patterns using a single NVIDIA RTX P40 GPU.}
\vspace{-5pt}
\small 
\setlength{\tabcolsep}{4pt} 
\renewcommand{\arraystretch}{1} 
\resizebox{0.7\linewidth}{!}{
\begin{tabular}{lcccccc}
\toprule
\textbf{Model} & \textbf{MDE} & \textbf{MDE Params.} & \textbf{Add. Params.} & \textbf{Latency (s)} & \textbf{AbsREL ($\downarrow$)} \\
\midrule
Marigold-DC & Marigold & 1.29B & 0M & 112.47 & 0.016 \\
TestPromptDC & UniDepth & 363.21M & 0.9M & 92.40 & 0.013 \\
PriorDA & DA-L & 335.3M & 96.4M & 0.68 & 0.011 \\
\midrule
Depth Anything & DA-L & 335.3M & 0M & 0.48 & 0.038 \\
\midrule
\multirow{3}{*}{Ours} 
 & DA-L & 335.3M & 60.6M & \cellcolor{bronze}0.49 & \best{0.007} \\
 & DA-B & 97.5M & 34.2M & \cellcolor{silver}\second{0.21} & \second{0.008} \\
 & DA-S & 24.8M & 8.9M & \cellcolor{gold}\best{0.09} & 0.009 \\
\bottomrule
\end{tabular}}
\label{tab:depth_comparison}
\vspace{-12pt}

\end{table}

Benefiting from this high efficiency and zero-shot accuracy, \textbf{Any2Full has been deployed in a real-world robotic warehouse} cell handling thousands of packages daily. It improved the grasping success rate for challenging black packages from 28\% to 91.6\% without damage (see Supp.~\ref{subsec:application}).

\begin{table*}[t]
\centering
\scriptsize
\renewcommand{\arraystretch}{1.03}
\caption{\textbf{Cross-backbone generalization of SAPE}. AbsREL ($\downarrow$) is evaluated across six benchmarks and multiple depth patterns.
``Any2Full (MoGe-2-B)'' denotes our model using MoGe-2-Base as the MDE backbone, while ``Any2Full (DA-B)'' denotes our model using Depth Anything-Base.
}
\vspace{-5pt}
\label{tab:cross_backbone}
\resizebox{\textwidth}{!}{
\begin{tabular}{l ccc ccc ccc ccc ccc ccc c}
\toprule
\multirow{2}{*}{Method}
& \multicolumn{3}{c}{\textbf{Average}}
& \multicolumn{3}{c}{NYU-Depth V2}
& \multicolumn{3}{c}{IBims-1}
& \multicolumn{3}{c}{DIODE}
& \multicolumn{3}{c}{KITTI DC}
& \multicolumn{3}{c}{ETH3D}
& \multicolumn{1}{c}{Logistic-Black}
\\
\cmidrule(lr){2-4}
\cmidrule(lr){5-7}
\cmidrule(lr){8-10}
\cmidrule(lr){11-13}
\cmidrule(lr){14-16}
\cmidrule(lr){17-19}
\cmidrule(lr){20-20}
& Hole & Range & Sparse
& Hole & Range & Random
& Hole & Range & Random
& Hole & Range & Random
& Hole & Range & LiDAR
& Hole & Range & SfM
& Mixed
\\
\midrule

MoGe-2-B
& 0.059 & 0.066 & 0.084
& 0.060 & 0.072 & 0.061
& 0.036 & 0.044 & 0.036
& 0.085 & 0.087 & 0.086
& 0.067 & 0.072 & 0.079
& 0.049 & 0.052 & 0.158
& 0.030
\\

Any2Full (MoGe-2-B)
& 0.012 & 0.051 & 0.034
& 0.010 & 0.043 & 0.018
& 0.009 & 0.038 & 0.014
& 0.020 & 0.066 & 0.025
& 0.015 & 0.060 & 0.021
& 0.008 & 0.048 & 0.090
& 0.011
\\
\textcolor{gray}{Any2Full (DA-B)}
& \textcolor{gray}{0.010} & \textcolor{gray}{0.046} & \textcolor{gray}{0.028}
& \textcolor{gray}{0.008} & \textcolor{gray}{0.035} & \textcolor{gray}{0.017}
& \textcolor{gray}{0.008} & \textcolor{gray}{0.031} & \textcolor{gray}{0.013}
& \textcolor{gray}{0.018} & \textcolor{gray}{0.063} & \textcolor{gray}{0.022}
& \textcolor{gray}{0.011} & \textcolor{gray}{0.058} & \textcolor{gray}{0.017}
& \textcolor{gray}{0.005} & \textcolor{gray}{0.041} & \textcolor{gray}{0.069}
& \textcolor{gray}{0.009}

\\

\bottomrule
\end{tabular}
}
\end{table*}

\vspace{4pt}
\noindent\textbf{Backbone Consistency.} 
We further train our model with different backbone sizes (Depth Anything-Large, Base, and Small) to examine the effectiveness of our approach, as shown in \cref{tab:zero-dc-quan}. Interestingly, a larger DA-L backbone does not necessarily improve performance, likely due to the coarse ground truth in datasets like NYU and KITTI, which limits the evaluation of richer geometric priors encoded in DA-L~\cite{yang2024depthanythingv2}. Nevertheless, our prompting strategy effectively provides strong scale priors that narrow the performance gap between backbones of different capacities, enabling consistent metric prediction across MDE sizes.

Beyond the Depth Anything family, we also conduct experiments using ViT-based MoGe-2-Base~\cite{wang2026moge} as the MDE backbone to verify the generality of our scale-prompting design. \cref{tab:cross_backbone} shows that SAPE can be effectively applied to MoGe-2 without architectural modification, consistently improving over the original MoGe-2 across all evaluated datasets and depth patterns. This demonstrates that our scale-prompting design is not specific to Depth Anything.

\vspace{4pt}
\noindent\textbf{Robustness to Depth Sparsity}. To evaluate robustness under varying depth sparsity levels, we conduct experiments on VIO-based sparse points (VOID) and LiDAR scans (KITTI) following standard sparsity protocols. As shown in \cref{tab:robustness_sparsity}, our method exhibits remarkable stability across diverse density levels. Notably, despite not being explicitly trained on sparse LiDAR data, both our method and TestPromptDC achieve competitive results on 4-line LiDAR scans compared to OMNI-DC. This underscores the benefits derived from the strong inherent geometric priors of MDE. Furthermore, while TestPromptDC suffers from significant performance degradation in extremely sparse VIO settings (e.g., VOID-150) due to unstable test-time adaptation, our approach maintains high precision without performance collapse, further confirming its superior robustness in depth completion tasks.
\begin{table}[t]
    \centering
    \vspace{-2pt}
    \caption{\textbf{Robustness comparison on varying depth sparsity levels.} We evaluate on the VOID dataset with 1500/500/150 sparse points sampled from a VIO system, and the KITTI dataset with 64/16/4-Line LiDAR scans.}
    \vspace{-5pt}
    \label{tab:robustness_sparsity}
    \small 
    \setlength{\tabcolsep}{4pt} 
    \resizebox{1\linewidth}{!}{
    \begin{tabular}{l cc !{\vrule width 0.8pt} cc cc cc cc cc cc}
        \toprule
        \multirow{2.5}{*}{\textbf{Method}} & \multicolumn{2}{c}{\textbf{AVG}} & \multicolumn{2}{c}{\textbf{VOID-1500}} & \multicolumn{2}{c}{\textbf{VOID-500}} & \multicolumn{2}{c}{\textbf{VOID-150}} & \multicolumn{2}{c}{\textbf{KITTI-64L}} & \multicolumn{2}{c}{\textbf{KITTI-16L}} & \multicolumn{2}{c}{\textbf{KITTI-4L}} \\
        \cmidrule(lr){2-3} \cmidrule(lr){4-5} \cmidrule(lr){6-7} \cmidrule(lr){8-9} \cmidrule(lr){10-11} \cmidrule(lr){12-13} \cmidrule(lr){14-15}
        & AbsREL & RMSE & AbsREL & RMSE & AbsREL & RMSE & AbsREL & RMSE & AbsREL & RMSE & AbsREL & RMSE & AbsREL & RMSE \\
        \midrule
        DepthPrompt & 0.122 & 3.187 & 0.046 & 0.623 & 0.080 & 0.648 & 0.115 & 0.755 & 0.098 & 5.005 & 0.106 & 4.701 & 0.283 & 7.391 \\
        
        OMNI-DC & \underline{0.032} & \underline{1.220} & \textbf{0.026} & \textbf{0.555} & \underline{0.039} & \underline{0.551} & \underline{0.057} & \underline{0.650} & \textbf{0.013} & \textbf{1.230} & \textbf{0.020} & \underline{1.703} & 0.039 & 2.630 \\
        
        TestPromptDC & 0.172 & 1.638 & 0.039 & 0.683 & 0.068 & 0.655 & 0.850 & 3.218 & 0.020 & \underline{1.350} & 0.023 & \textbf{1.699} & \textbf{0.032} & \textbf{2.221} \\
        
        \midrule
        \rowcolor{oursrow}
        Ours & \textbf{0.031} & \textbf{1.203} & \underline{0.030} & \underline{0.561} & \textbf{0.037} & \textbf{0.538} & \textbf{0.045} & \textbf{0.620} & \underline{0.017} & 1.356 & \underline{0.023} & 1.760 & \underline{0.035} & \underline{2.380} \\
        \bottomrule
    \end{tabular}}

\vspace{-5pt}
\end{table}

\begin{figure}[t]

    \vspace{-5pt}
    \centering
    \includegraphics[width=0.6\linewidth]{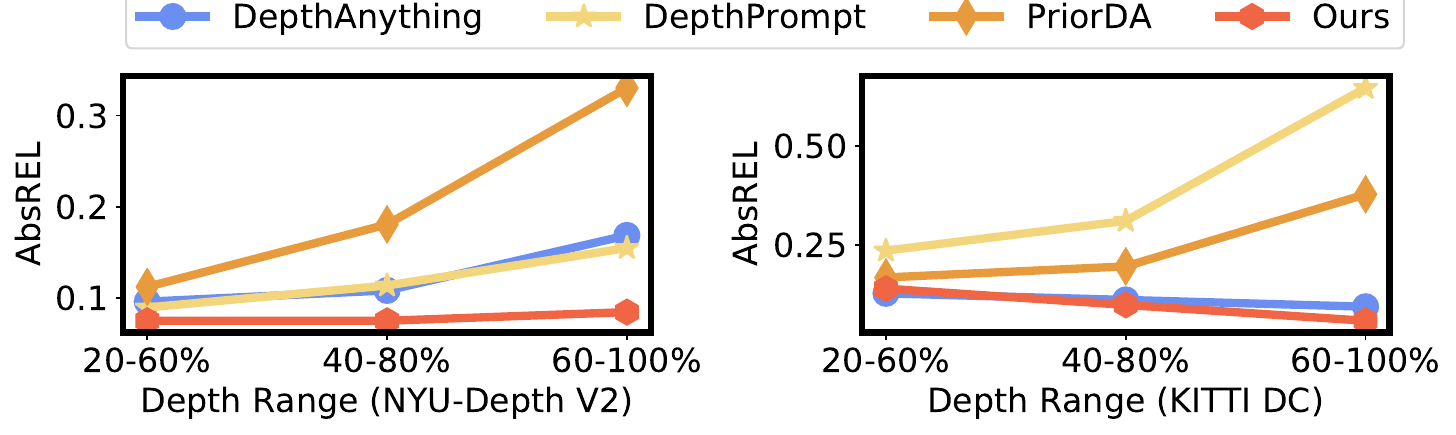}
    \vspace{-5pt}
    \caption{\textbf{Robustness comparison on varying depth range.} Evaluation is conducted using AbsREL ($\downarrow$) on NYU-Depth V2 and KITTI DC. }
    \label{fig:robustness-range}
    \vspace{-5pt}
\end{figure}

\vspace{4pt}
\noindent\textbf{Robustness to Depth Range}. To evaluate robustness under varying depth range, we conduct experiments on the indoor NYU-Depth V2 and outdoor KITTI datasets using different valid depth ranges. As shown in \cref{fig:robustness-range}, our method Any2Full exhibits remarkable consistency across all depth intervals. In contrast, existing MDE-based feed-forward methods, such as DepthPrompt and PriorDA, suffer from significant performance degradation when valid depths are confined to narrow ranges (e.g., the 60-100\% interval on KITTI). This highlights a fundamental challenge in scale extrapolation when global range information is limited. By leveraging MDE-guided geometric reasoning to propagate local scale cues, Any2Full effectively extrapolates globally consistent scales to unmeasured regions, ensuring robust completion even under constrained depth observation.

\vspace{-2pt}
\subsection{Ablations and Analysis}

\begin{table}[t]
\centering
\caption{
\textbf{Ablation study} of Any2Full based on DA-B. 
\textbf{SP}: scale prompt fusion, \textbf{LE}: local enrichment, \textbf{GP}: global propagation. 
Metric is AbsREL ($\downarrow$) evaluated on IBims-1 and KITTI DC across different depth patterns and sparsity levels.
}
\vspace{-5pt}
\small 
\setlength{\tabcolsep}{4pt} 
\renewcommand{\arraystretch}{1} 
\resizebox{1\linewidth}{!}{ 

\begin{tabular}{lccccccccc | cccccc}
\toprule
\multirow{2}{*}{\textbf{Model}} & \multirow{2}{*}{\textbf{SP}} & \multirow{2}{*}{\textbf{LE}} & \multirow{2}{*}{\textbf{GP}} &
\multicolumn{3}{c}{\textbf{IBims-1 }} & 
\multicolumn{3}{c}{\textbf{KITTI DC}} & 
\multicolumn{3}{c}{\textbf{IBims-1 }} & 
\multicolumn{3}{c}{\textbf{KITTI DC}}\\
& & & & Hole & Range & Sparse & Hole & Range & Sparse &
 10\% & 1\% & 0.1\% & 64L & 16L & 4L \\
\midrule
DA-B &  &  &  & 0.047 & 0.048 & 0.047 & 0.079 & 0.087 & 0.079 
& 0.044 & 0.044 &  0.044 & 0.086 & 0.087 & 0.089\\
\midrule
Any2Full & \checkmark &  &  & 0.010 & 0.068 & 0.022 & 0.017 & \second{0.089} & \second{0.039} 
& 0.011 & 0.020 & \second{0.036} &  0.021 &  0.032 & 0.062\\
Any2Full & \checkmark & \checkmark &  & \second{0.009} & \second{0.066} & \second{0.020} & \second{0.015} & 0.092 & 0.044 
& \second{0.011} & \second{0.017} & 0.039 &  \second{0.021} &  \second{0.032} & 0.065\\
Any2Full & \checkmark & \checkmark & \checkmark & \best{0.008} & \best{0.031} & \best{0.013} & \best{0.011} & \best{0.046} & \best{0.021} 
& \best{0.010} & \best{0.012} & \best{0.018} &  \best{0.017} &  \best{0.022} & \best{0.034}\\
\bottomrule
\end{tabular}}

\label{tab:ablation}
\vspace{-5pt}
\end{table}

To assess the contribution of each module, we conduct ablation studies in \cref{tab:ablation} by progressively removing Local Enrichment (LE) and Global Propagation (GP) modules.
Comparing Any2Full w/o LE\&GP with the DA-B baseline, we observe that the scale prompt (SP) fusion effectively provides necessary scale cues, enabling the MDE backbone to adapt for accurate metric prediction across domains.
Furthermore, incorporating the LE and GP modules consistently improves accuracy under varying depth patterns and sparsity levels.
Detailed module analyses and additional ablations are provided in the Supp.~\ref{supsec:ablation}.



\vspace{-4pt}
\section{Conclusion}
\vspace{-4pt}
In this paper, we present Any2Full, a domain-general and pattern-agnostic depth completion model in one stage. Unlike traditional RGB-D fusion methods, Any2Full reformulates depth completion as a scale-prompting adaptation of monocular depth estimation. It inherits the domain-general geometric priors of a pretrained MDE backbone while flexibly adapting to diverse sparse depth patterns through a scale-aware prompt encoder. Any2Full achieves superior generalization, speed, and accuracy, with practical effectiveness demonstrated in real-world deployment, offering a new perspective on efficient and unified depth completion.
\label{sec:conclusion}

\newpage
%
%

\bibliographystyle{splncs04}
\bibliography{main}

@String(ECCV  = {Eur. Conf. Comput. Vis.})

@String(AAAI  = {AAAI})

@String(ECCV  = {ECCV})

@inproceedings{zhang2023completionformer,
  title={Completionformer: Depth completion with convolutions and vision transformers},
  author={Zhang, Youmin and Guo, Xianda and Poggi, Matteo and Zhu, Zheng and Huang, Guan and Mattoccia, Stefano},
  booktitle={Proceedings of the IEEE/CVF conference on computer vision and pattern recognition},
  pages={18527--18536},
  year={2023}
}

@inproceedings{park2024depthprompt,
  title={Depth prompting for sensor-agnostic depth estimation},
  author={Park, Jin-Hwi and Jeong, Chanhwi and Lee, Junoh and Jeon, Hae-Gon},
  booktitle={Proceedings of the IEEE/CVF Conference on Computer Vision and Pattern Recognition},
  pages={9859--9869},
  year={2024}
}

@inproceedings{ke2024marigold,
  title={Repurposing diffusion-based image generators for monocular depth estimation},
  author={Ke, Bingxin and Obukhov, Anton and Huang, Shengyu and Metzger, Nando and Daudt, Rodrigo Caye and Schindler, Konrad},
  booktitle={Proceedings of the IEEE/CVF conference on computer vision and pattern recognition},
  pages={9492--9502},
  year={2024}
}

@inproceedings{viola2025marigolddc,
  title={Marigold-dc: Zero-shot monocular depth completion with guided diffusion},
  author={Viola, Massimiliano and Qu, Kevin and Metzger, Nando and Ke, Bingxin and Becker, Alexander and Schindler, Konrad and Obukhov, Anton},
  booktitle={Proceedings of the IEEE/CVF International Conference on Computer Vision},
  pages={5359--5370},
  year={2025}
}

@inproceedings{yang2024depthanythingv1,
  title={Depth anything: Unleashing the power of large-scale unlabeled data},
  author={Yang, Lihe and Kang, Bingyi and Huang, Zilong and Xu, Xiaogang and Feng, Jiashi and Zhao, Hengshuang},
  booktitle={Proceedings of the IEEE/CVF conference on computer vision and pattern recognition},
  pages={10371--10381},
  year={2024}
}

@article{yang2024depthanythingv2,
  title={Depth anything v2},
  author={Yang, Lihe and Kang, Bingyi and Huang, Zilong and Zhao, Zhen and Xu, Xiaogang and Feng, Jiashi and Zhao, Hengshuang},
  journal={Advances in Neural Information Processing Systems},
  volume={37},
  pages={21875--21911},
  year={2024}
}

@inproceedings{lin2025promptingda,
  title={Prompting depth anything for 4k resolution accurate metric depth estimation},
  author={Lin, Haotong and Peng, Sida and Chen, Jingxiao and Peng, Songyou and Sun, Jiaming and Liu, Minghuan and Bao, Hujun and Feng, Jiashi and Zhou, Xiaowei and Kang, Bingyi},
  booktitle={Proceedings of the Computer Vision and Pattern Recognition Conference},
  pages={17070--17080},
  year={2025}
}

@article{wang2025priorda,
  title={Depth Anything with Any Prior},
  author={Wang, Zehan and Chen, Siyu and Yang, Lihe and Wang, Jialei and Zhang, Ziang and Zhao, Hengshuang and Zhao, Zhou},
  journal={arXiv preprint arXiv:2505.10565},
  year={2025}
}

@inproceedings{jeong2025testprompt,
  title={Test-Time Prompt Tuning for Zero-Shot Depth Completion},
  author={Jeong, Chanhwi and Bae, Inhwan and Park, Jin-Hwi and Jeon, Hae-Gon},
  booktitle={Proceedings of the IEEE/CVF International Conference on Computer Vision},
  pages={9443--9454},
  year={2025}
}

@article{arampatzakis2023monocularrevier,
  title={Monocular depth estimation: A thorough review},
  author={Arampatzakis, Vasileios and Pavlidis, George and Mitianoudis, Nikolaos and Papamarkos, Nikos},
  journal={IEEE Transactions on Pattern Analysis and Machine Intelligence},
  volume={46},
  number={4},
  pages={2396--2414},
  year={2023},
  publisher={IEEE}
}

@article{eigen2014depth,
  title={Depth map prediction from a single image using a multi-scale deep network},
  author={Eigen, David and Puhrsch, Christian and Fergus, Rob},
  journal={Advances in neural information processing systems},
  volume={27},
  year={2014}
}

@article{brown2020language,
  title={Language models are few-shot learners},
  author={Brown, Tom and Mann, Benjamin and Ryder, Nick and Subbiah, Melanie and Kaplan, Jared D and Dhariwal, Prafulla and Neelakantan, Arvind and Shyam, Pranav and Sastry, Girish and Askell, Amanda and others},
  journal={Advances in neural information processing systems},
  volume={33},
  pages={1877--1901},
  year={2020}
}

@inproceedings{devlin2019bert,
  title={Bert: Pre-training of deep bidirectional transformers for language understanding},
  author={Devlin, Jacob and Chang, Ming-Wei and Lee, Kenton and Toutanova, Kristina},
  booktitle={Proceedings of the 2019 conference of the North American chapter of the association for computational linguistics: human language technologies, volume 1 (long and short papers)},
  pages={4171--4186},
  year={2019}
}

@article{ranftl2020towards,
  title={Towards robust monocular depth estimation: Mixing datasets for zero-shot cross-dataset transfer},
  author={Ranftl, Ren{\'e} and Lasinger, Katrin and Hafner, David and Schindler, Konrad and Koltun, Vladlen},
  journal={IEEE transactions on pattern analysis and machine intelligence},
  volume={44},
  number={3},
  pages={1623--1637},
  year={2020},
  publisher={IEEE}
}

@inproceedings{moon2024ground,
  title={From-ground-to-objects: Coarse-to-fine self-supervised monocular depth estimation of dynamic objects with ground contact prior},
  author={Moon, Jaeho and Bello, Juan Luis Gonzalez and Kwon, Byeongjun and Kim, Munchurl},
  booktitle={Proceedings of the IEEE/CVF Conference on Computer Vision and Pattern Recognition},
  pages={10519--10529},
  year={2024}
}

@inproceedings{gui2025depthfm,
  title={Depthfm: Fast generative monocular depth estimation with flow matching},
  author={Gui, Ming and Schusterbauer, Johannes and Prestel, Ulrich and Ma, Pingchuan and Kotovenko, Dmytro and Grebenkova, Olga and Baumann, Stefan Andreas and Hu, Vincent Tao and Ommer, Bj{\"o}rn},
  booktitle={Proceedings of the AAAI Conference on Artificial Intelligence},
  pages={3203--3211},
  year={2025}
}

@inproceedings{wang2025moge,
  title={Moge: Unlocking accurate monocular geometry estimation for open-domain images with optimal training supervision},
  author={Wang, Ruicheng and Xu, Sicheng and Dai, Cassie and Xiang, Jianfeng and Deng, Yu and Tong, Xin and Yang, Jiaolong},
  booktitle={Proceedings of the IEEE/CVF Conference on Computer Vision and Pattern Recognition},
  pages={5261--5271},
  year={2025}
}

@article{wang2026moge,
  title={Moge-2: Accurate monocular geometry with metric scale and sharp details},
  author={Wang, Ruicheng and Xu, Sicheng and Dong, Yue and Deng, Yu and Xiang, Jianfeng and Lv, Zelong and Sun, Guangzhong and Tong, Xin and Yang, Jiaolong},
  journal={Advances in Neural Information Processing Systems},
  volume={38},
  pages={35928--35959},
  year={2026}
}

@inproceedings{zeng2024wordepth,
  title={Wordepth: Variational language prior for monocular depth estimation},
  author={Zeng, Ziyao and Wang, Daniel and Yang, Fengyu and Park, Hyoungseob and Soatto, Stefano and Lao, Dong and Wong, Alex},
  booktitle={Proceedings of the IEEE/CVF Conference on Computer Vision and Pattern Recognition},
  pages={9708--9719},
  year={2024}
}

@article{bhat2023zoedepth,
  title={Zoedepth: Zero-shot transfer by combining relative and metric depth},
  author={Bhat, Shariq Farooq and Birkl, Reiner and Wofk, Diana and Wonka, Peter and M{\"u}ller, Matthias},
  journal={arXiv preprint arXiv:2302.12288},
  year={2023}
}

@inproceedings{yin2023metric3d,
  title={Metric3d: Towards zero-shot metric 3d prediction from a single image},
  author={Yin, Wei and Zhang, Chi and Chen, Hao and Cai, Zhipeng and Yu, Gang and Wang, Kaixuan and Chen, Xiaozhi and Shen, Chunhua},
  booktitle={Proceedings of the IEEE/CVF international conference on computer vision},
  pages={9043--9053},
  year={2023}
}

@inproceedings{piccinelli2024unidepth,
  title={UniDepth: Universal monocular metric depth estimation},
  author={Piccinelli, Luigi and Yang, Yung-Hsu and Sakaridis, Christos and Segu, Mattia and Li, Siyuan and Van Gool, Luc and Yu, Fisher},
  booktitle={Proceedings of the IEEE/CVF Conference on Computer Vision and Pattern Recognition},
  pages={10106--10116},
  year={2024}
}

@article{piccinelli2025unidepthv2,
  title={Unidepthv2: Universal monocular metric depth estimation made simpler},
  author={Piccinelli, Luigi and Sakaridis, Christos and Yang, Yung-Hsu and Segu, Mattia and Li, Siyuan and Abbeloos, Wim and Van Gool, Luc},
  journal={IEEE Transactions on Pattern Analysis and Machine Intelligence},
  year={2025},
  publisher={IEEE}
}

@article{bochkovskii2024depth,
  title={Depth pro: Sharp monocular metric depth in less than a second},
  author={Bochkovskii, Aleksei and Delaunoy, Ama{\~A}{\c{G}}l and Germain, Hugo and Santos, Marcel and Zhou, Yichao and Richter, Stephan R and Koltun, Vladlen},
  journal={arXiv preprint arXiv:2410.02073},
  year={2024}
}

@article{zhu2024scaledepth,
  title={Scaledepth: Decomposing metric depth estimation into scale prediction and relative depth estimation},
  author={Zhu, Ruijie and Wang, Chuxin and Song, Ziyang and Liu, Li and Zhang, Tianzhu and Zhang, Yongdong},
  journal={arXiv preprint arXiv:2407.08187},
  year={2024}
}

@inproceedings{ma2018sparse,
  title={Sparse-to-dense: Depth prediction from sparse depth samples and a single image},
  author={Ma, Fangchang and Karaman, Sertac},
  booktitle={2018 IEEE international conference on robotics and automation (ICRA)},
  pages={4796--4803},
  year={2018},
  organization={IEEE}
}

@article{tang2020learning,
  title={Learning guided convolutional network for depth completion},
  author={Tang, Jie and Tian, Fei-Peng and Feng, Wei and Li, Jian and Tan, Ping},
  journal={IEEE Transactions on Image Processing},
  volume={30},
  pages={1116--1129},
  year={2020},
  publisher={IEEE}
}

@inproceedings{rho2022guideformer,
  title={Guideformer: Transformers for image guided depth completion},
  author={Rho, Kyeongha and Ha, Jinsung and Kim, Youngjung},
  booktitle={Proceedings of the IEEE/CVF Conference on Computer Vision and Pattern Recognition},
  pages={6250--6259},
  year={2022}
}

@inproceedings{jun2024masked,
  title={Masked spatial propagation network for sparsity-adaptive depth refinement},
  author={Jun, Jinyoung and Lee, Jae-Han and Kim, Chang-Su},
  booktitle={Proceedings of the IEEE/CVF Conference on Computer Vision and Pattern Recognition},
  pages={19768--19778},
  year={2024}
}

@inproceedings{cheng2018depthcspn,
  title={Depth estimation via affinity learned with convolutional spatial propagation network},
  author={Cheng, Xinjing and Wang, Peng and Yang, Ruigang},
  booktitle={Proceedings of the European conference on computer vision (ECCV)},
  pages={103--119},
  year={2018}
}

@article{cheng2019learning,
  title={Learning depth with convolutional spatial propagation network},
  author={Cheng, Xinjing and Wang, Peng and Yang, Ruigang},
  journal={IEEE transactions on pattern analysis and machine intelligence},
  volume={42},
  number={10},
  pages={2361--2379},
  year={2019},
  publisher={IEEE}
}

@inproceedings{cheng2020cspn++,
  title={Cspn++: Learning context and resource aware convolutional spatial propagation networks for depth completion},
  author={Cheng, Xinjing and Wang, Peng and Guan, Chenye and Yang, Ruigang},
  booktitle={Proceedings of the AAAI conference on artificial intelligence},
  volume={34},
  pages={10615--10622},
  year={2020}
}

@inproceedings{park2020nonlocalspn,
  title={Non-local spatial propagation network for depth completion},
  author={Park, Jinsun and Joo, Kyungdon and Hu, Zhe and Liu, Chi-Kuei and So Kweon, In},
  booktitle={European conference on computer vision},
  pages={120--136},
  year={2020},
  organization={Springer}
}

@inproceedings{lin2022dynamic,
  title={Dynamic spatial propagation network for depth completion},
  author={Lin, Yuankai and Cheng, Tao and Zhong, Qi and Zhou, Wending and Yang, Hua},
  booktitle={Proceedings of the aaai conference on artificial intelligence},
  volume={36},
  pages={1638--1646},
  year={2022}
}

@inproceedings{liu2022graphcspn,
  title={Graphcspn: Geometry-aware depth completion via dynamic gcns},
  author={Liu, Xin and Shao, Xiaofei and Wang, Bo and Li, Yali and Wang, Shengjin},
  booktitle={European Conference on Computer Vision},
  pages={90--107},
  year={2022},
  organization={Springer}
}

@inproceedings{tang2024bispn,
  title={Bilateral propagation network for depth completion},
  author={Tang, Jie and Tian, Fei-Peng and An, Boshi and Li, Jian and Tan, Ping},
  booktitle={Proceedings of the IEEE/CVF Conference on Computer Vision and Pattern Recognition},
  pages={9763--9772},
  year={2024}
}

@inproceedings{wang2022depthhole,
  title={Rgb-depth fusion gan for indoor depth completion},
  author={Wang, Haowen and Wang, Mingyuan and Che, Zhengping and Xu, Zhiyuan and Qiao, Xiuquan and Qi, Mengshi and Feng, Feifei and Tang, Jian},
  booktitle={Proceedings of the ieee/cvf conference on computer vision and pattern recognition},
  pages={6209--6218},
  year={2022}
}

@article{wang2024depthholev2,
  title={Rdfc-gan: Rgb-depth fusion cyclegan for indoor depth completion},
  author={Wang, Haowen and Che, Zhengping and Yang, Yufan and Wang, Mingyuan and Xu, Zhiyuan and Qiao, Xiuquan and Qi, Mengshi and Feng, Feifei and Tang, Jian},
  journal={IEEE Transactions on Pattern Analysis and Machine Intelligence},
  volume={46},
  number={11},
  pages={7088--7101},
  year={2024},
  publisher={IEEE}
}

@inproceedings{hyoseok2025zero,
  title={Zero-shot Depth Completion via Test-time Alignment with Affine-invariant Depth Prior},
  author={Hyoseok, Lee and Kim, Kyeong Seon and Byung-Ki, Kwon and Oh, Tae-Hyun},
  booktitle={Proceedings of the AAAI Conference on Artificial Intelligence},
  volume={39},
  pages={3877--3885},
  year={2025}
}

@inproceedings{jeong2025test,
  title={Test-Time Prompt Tuning for Zero-Shot Depth Completion},
  author={Jeong, Chanhwi and Bae, Inhwan and Park, Jin-Hwi and Jeon, Hae-Gon},
  booktitle={Proceedings of the IEEE/CVF International Conference on Computer Vision},
  pages={9443--9454},
  year={2025}
}

@inproceedings{yularge2026,
  title={Large Depth Completion Model from Sparse Observations},
  author={Yu, Zhu and Zhang, Runmin and Qiu, Lingteng and Cao, Si-Yuan and Qiu, Kejie and He, Yisheng and Zhu, Siyu and Dong, Zilong and Shen, Hui-liang and others},
  booktitle={The Fourteenth International Conference on Learning Representations},
  year={2026}
}

@article{miao2023occdepth,
  title={Occdepth: A depth-aware method for 3d semantic scene completion},
  author={Miao, Ruihang and Liu, Weizhou and Chen, Mingrui and Gong, Zheng and Xu, Weixin and Hu, Chen and Zhou, Shuchang},
  journal={arXiv preprint arXiv:2302.13540},
  year={2023}
}

@inproceedings{wang2025tacodepth,
  title={TacoDepth: Towards Efficient Radar-Camera Depth Estimation with One-stage Fusion},
  author={Wang, Yiran and Li, Jiaqi and Hong, Chaoyi and Li, Ruibo and Sun, Liusheng and Song, Xiao and Wang, Zhe and Cao, Zhiguo and Lin, Guosheng},
  booktitle={Proceedings of the Computer Vision and Pattern Recognition Conference},
  pages={10523--10533},
  year={2025}
}

@inproceedings{cheng2025monster,
  title={Monster: Marry monodepth to stereo unleashes power},
  author={Cheng, Junda and Liu, Longliang and Xu, Gangwei and Wang, Xianqi and Zhang, Zhaoxing and Deng, Yong and Zang, Jinliang and Chen, Yurui and Cai, Zhipeng and Yang, Xin},
  booktitle={Proceedings of the Computer Vision and Pattern Recognition Conference},
  pages={6273--6282},
  year={2025}
}

@inproceedings{zuo2025omni,
  title={Omni-dc: Highly robust depth completion with multiresolution depth integration},
  author={Zuo, Yiming and Yang, Willow and Ma, Zeyu and Deng, Jia},
  booktitle={Proceedings of the IEEE/CVF International Conference on Computer Vision},
  pages={9287--9297},
  year={2025}
}

@inproceedings{wang2025pacgdc,
  title={PacGDC: Label-Efficient Generalizable Depth Completion with Projection Ambiguity and Consistency},
  author={Wang, Haotian and Xiao, Aoran and Zhang, Xiaoqin and Yang, Meng and Lu, Shijian},
  booktitle={Proceedings of the IEEE/CVF International Conference on Computer Vision},
  pages={7709--7720},
  year={2025}
}

@article{wang2023g2,
  title={G2-monodepth: A general framework of generalized depth inference from monocular rgb+ x data},
  author={Wang, Haotian and Yang, Meng and Zheng, Nanning},
  journal={IEEE Transactions on Pattern Analysis and Machine Intelligence},
  volume={46},
  number={5},
  pages={3753--3771},
  year={2023},
  publisher={IEEE}
}

@article{ma2026metricanything,
  title={MetricAnything: Scaling Metric Depth Pretraining with Noisy Heterogeneous Sources},
  author={Ma, Baorui and Yang, Jiahui and Di, Donglin and Zhang, Xuancheng and Cui, Jianxun and Li, Hao and Xie, Yan and Chen, Wei},
  journal={arXiv preprint arXiv:2601.22054},
  year={2026}
}

@inproceedings{roberts2021hypersim,
  title={Hypersim: A photorealistic synthetic dataset for holistic indoor scene understanding},
  author={Roberts, Mike and Ramapuram, Jason and Ranjan, Anurag and Kumar, Atulit and Bautista, Miguel Angel and Paczan, Nathan and Webb, Russ and Susskind, Joshua M},
  booktitle={Proceedings of the IEEE/CVF international conference on computer vision},
  pages={10912--10922},
  year={2021}
}

@article{cabon2020vkitti,
  title={Virtual kitti 2},
  author={Cabon, Yohann and Murray, Naila and Humenberger, Martin},
  journal={arXiv preprint arXiv:2001.10773},
  year={2020}
}

@inproceedings{wang2020tartanair,
  title={Tartanair: A dataset to push the limits of visual slam},
  author={Wang, Wenshan and Zhu, Delong and Wang, Xiangwei and Hu, Yaoyu and Qiu, Yuheng and Wang, Chen and Hu, Yafei and Kapoor, Ashish and Scherer, Sebastian},
  booktitle={2020 IEEE/RSJ International Conference on Intelligent Robots and Systems (IROS)},
  pages={4909--4916},
  year={2020},
  organization={IEEE}
}

@inproceedings{silberman2012nyuv2,
  title={Indoor segmentation and support inference from rgbd images},
  author={Silberman, Nathan and Hoiem, Derek and Kohli, Pushmeet and Fergus, Rob},
  booktitle={European conference on computer vision},
  pages={746--760},
  year={2012},
  organization={Springer}
}

@inproceedings{koch2018ibims,
  title={Evaluation of cnn-based single-image depth estimation methods},
  author={Koch, Tobias and Liebel, Lukas and Fraundorfer, Friedrich and Korner, Marco},
  booktitle={Proceedings of the European Conference on Computer Vision (ECCV) Workshops},
  pages={0--0},
  year={2018}
}

@inproceedings{uhrig2017kittidc,
  title={Sparsity invariant cnns},
  author={Uhrig, Jonas and Schneider, Nick and Schneider, Lukas and Franke, Uwe and Brox, Thomas and Geiger, Andreas},
  booktitle={2017 international conference on 3D Vision (3DV)},
  pages={11--20},
  year={2017},
  organization={IEEE}
}

@article{vasiljevic2019diode,
  title={Diode: A dense indoor and outdoor depth dataset},
  author={Vasiljevic, Igor and Kolkin, Nick and Zhang, Shanyi and Luo, Ruotian and Wang, Haochen and Dai, Falcon Z and Daniele, Andrea F and Mostajabi, Mohammadreza and Basart, Steven and Walter, Matthew R and others},
  journal={arXiv preprint arXiv:1908.00463},
  year={2019}
}

@inproceedings{schops2017eth3d,
  title={A multi-view stereo benchmark with high-resolution images and multi-camera videos},
  author={Schops, Thomas and Schonberger, Johannes L and Galliani, Silvano and Sattler, Torsten and Schindler, Konrad and Pollefeys, Marc and Geiger, Andreas},
  booktitle={Proceedings of the IEEE conference on computer vision and pattern recognition},
  pages={3260--3269},
  year={2017}
}

@article{wong2020unsupervised,
  title={Unsupervised depth completion from visual inertial odometry},
  author={Wong, Alex and Fei, Xiaohan and Tsuei, Stephanie and Soatto, Stefano},
  journal={IEEE Robotics and Automation Letters},
  volume={5},
  number={2},
  pages={1899--1906},
  year={2020},
  publisher={IEEE}
}

@article{tang2022perception,
  title={Perception and navigation in autonomous systems in the era of learning: A survey},
  author={Tang, Yang and Zhao, Chaoqiang and Wang, Jianrui and Zhang, Chongzhen and Sun, Qiyu and Zheng, Wei Xing and Du, Wenli and Qian, Feng and Kurths, Juergen},
  journal={IEEE Transactions on Neural Networks and Learning Systems},
  volume={34},
  number={12},
  pages={9604--9624},
  year={2022},
  publisher={IEEE}
}

@inproceedings{maier2012real,
  title={Real-time navigation in 3D environments based on depth camera data},
  author={Maier, Daniel and Hornung, Armin and Bennewitz, Maren},
  booktitle={2012 12th IEEE-RAS International Conference on Humanoid Robots (Humanoids 2012)},
  pages={692--697},
  year={2012},
  organization={IEEE}
}

@article{mahler2019learning,
  title={Learning ambidextrous robot grasping policies},
  author={Mahler, Jeffrey and Matl, Matthew and Satish, Vishal and Danielczuk, Michael and DeRose, Bill and McKinley, Stephen and Goldberg, Ken},
  journal={Science Robotics},
  volume={4},
  number={26},
  pages={eaau4984},
  year={2019},
  publisher={American Association for the Advancement of Science}
}

@article{du2021vision,
  title={Vision-based robotic grasping from object localization, object pose estimation to grasp estimation for parallel grippers: a review},
  author={Du, Guoguang and Wang, Kai and Lian, Shiguo and Zhao, Kaiyong},
  journal={Artificial Intelligence Review},
  volume={54},
  number={3},
  pages={1677--1734},
  year={2021},
  publisher={Springer}
}

@article{tan2024attention,
  title={Attention-based grasp detection with monocular depth estimation},
  author={Tan, Phan Xuan and Hoang, Dinh-Cuong and Nguyen, Anh-Nhat and Nguyen, Van-Thiep and Vu, Van-Duc and Nguyen, Thu-Uyen and Hoang, Ngoc-Anh and Phan, Khanh-Toan and Tran, Duc-Thanh and Vu, Duy-Quang and others},
  journal={IEEE Access},
  volume={12},
  pages={65041--65057},
  year={2024},
  publisher={IEEE}
}

@article{armeni2017joint,
  title={Joint 2d-3d-semantic data for indoor scene understanding},
  author={Armeni, Iro and Sax, Sasha and Zamir, Amir R and Savarese, Silvio},
  journal={arXiv preprint arXiv:1702.01105},
  year={2017}
}

@inproceedings{chen2019towards,
  title={Towards scene understanding: Unsupervised monocular depth estimation with semantic-aware representation},
  author={Chen, Po-Yi and Liu, Alexander H and Liu, Yen-Cheng and Wang, Yu-Chiang Frank},
  booktitle={Proceedings of the IEEE/CVF Conference on computer vision and pattern recognition},
  pages={2624--2632},
  year={2019}
}

@inproceedings{geiger2012we,
  title={Are we ready for autonomous driving? the kitti vision benchmark suite},
  author={Geiger, Andreas and Lenz, Philip and Urtasun, Raquel},
  booktitle={2012 IEEE conference on computer vision and pattern recognition},
  pages={3354--3361},
  year={2012},
  organization={IEEE}
}

@inproceedings{schonberger2016structure,
  title={Structure-from-motion revisited},
  author={Schonberger, Johannes L and Frahm, Jan-Michael},
  booktitle={Proceedings of the IEEE conference on computer vision and pattern recognition},
  pages={4104--4113},
  year={2016}
}

@article{foix2011lock,
  title={Lock-in time-of-flight (ToF) cameras: A survey},
  author={Foix, Sergi and Alenya, Guillem and Torras, Carme},
  journal={IEEE Sensors Journal},
  volume={11},
  number={9},
  pages={1917--1926},
  year={2011},
  publisher={IEEE}
}

@article{dosovitskiy2020image,
  title={An image is worth 16x16 words: Transformers for image recognition at scale},
  author={Dosovitskiy, Alexey},
  journal={arXiv preprint arXiv:2010.11929},
  year={2020}
}

@inproceedings{perez2018film,
  title={FiLM: Visual reasoning with a general conditioning layer},
  author={Perez, Ethan and Strub, Florian and De Vries, Harm and Dumoulin, Vincent and Courville, Aaron},
  booktitle={Proceedings of the AAAI conference on artificial intelligence},
  volume={32},
  year={2018}
}

@inproceedings{jia2022visual,
  title={Visual prompt tuning},
  author={Jia, Menglin and Tang, Luming and Chen, Bor-Chun and Cardie, Claire and Belongie, Serge and Hariharan, Bharath and Lim, Ser-Nam},
  booktitle={European conference on computer vision},
  pages={709--727},
  year={2022},
  organization={Springer}
}

@article{li2021prefix,
  title={Prefix-tuning: Optimizing continuous prompts for generation},
  author={Li, Xiang Lisa and Liang, Percy},
  journal={arXiv preprint arXiv:2101.00190},
  year={2021}
}

@article{loshchilov2016sgdr,
  title={Sgdr: Stochastic gradient descent with warm restarts},
  author={Loshchilov, Ilya and Hutter, Frank},
  journal={arXiv preprint arXiv:1608.03983},
  year={2016}
}

@article{kingma2014adam,
  title={Adam: A method for stochastic optimization},
  author={Kingma, Diederik P},
  journal={arXiv preprint arXiv:1412.6980},
  year={2014}
}

@inproceedings{koch2018evaluation,
  title={Evaluation of cnn-based single-image depth estimation methods},
  author={Koch, Tobias and Liebel, Lukas and Fraundorfer, Friedrich and Korner, Marco},
  booktitle={Proceedings of the European Conference on Computer Vision (ECCV) Workshops},
  pages={0--0},
  year={2018}
}

@article{fischler1981random,
  title={Random sample consensus: a paradigm for model fitting with applications to image analysis and automated cartography},
  author={Fischler, Martin A and Bolles, Robert C},
  journal={Communications of the ACM},
  volume={24},
  number={6},
  pages={381--395},
  year={1981},
  publisher={ACM New York, NY, USA}
}

@inproceedings{xinyue,
author = {Feng, Xinyue and Zhong, Shuxin and Hang, Jinquan and Lyu, Wenjun and Zhang, Yuequn and Yang, Guang and Wang, Haotian and Zhang, Desheng and Wang, Guang},
title = {Hierarchical Structure Sharing Empowers Multi-task Heterogeneous GNNs for Customer Expansion},
year = {2025},
isbn = {9798400714542},
publisher = {Association for Computing Machinery},
address = {New York, NY, USA},
booktitle = {Proceedings of the 31st ACM SIGKDD Conference on Knowledge Discovery and Data Mining V.2},
pages = {4424–4434},
numpages = {11},
location = {Toronto ON, Canada},
series = {KDD '25}
}

@inproceedings{zejun,
  author       = {Zejun Xie and
                  Wenjun Lyu and
                  Yiwei Song and
                  Haotian Wang and
                  Guang Yang and
                  Yunhuai Liu and
                  Tian He and
                  Desheng Zhang and
                  Guang Wang},
  editor       = {Yizhou Sun and
                  Flavio Chierichetti and
                  Hady W. Lauw and
                  Claudia Perlich and
                  Wee Hyong Tok and
                  Andrew Tomkins},
  title        = {Scalable Area Difficulty Assessment with Knowledge-enhanced {AI} for
                  Nationwide Logistics Systems},
  booktitle    = {Proceedings of the 31st {ACM} {SIGKDD} Conference on Knowledge Discovery
                  and Data Mining, V.1, {KDD} 2025, Toronto, ON, Canada, August 3-7,
                  2025},
  pages        = {2713--2724},
  publisher    = {{ACM}},
  year         = {2025},
  bibsource    = {dblp computer science bibliography, https://dblp.org}
}

@inproceedings{zhiyuan,
author = {Zhou, Zhiyuan and Lin, Li and Wang, Hai and Zhou, Xiaolei and Wei, Gong and Wang, Shuai},
title = {A Cross Domain Method for Customer Lifetime Value Prediction in Supply Chain Platform},
year = {2024},
isbn = {9798400701719},
publisher = {Association for Computing Machinery},
address = {New York, NY, USA},
booktitle = {Proceedings of the ACM Web Conference 2024},
pages = {4037–4046},
numpages = {10},
location = {Singapore, Singapore},
series = {WWW '24}
}

\clearpage
\appendix
\setcounter{page}{1}
\section*{Overview}
This supplementary material provides additional results, implementation details, and analyses. It includes:

\begin{itemize}
    \item Additional experiments and analyses (Sec.~\ref{sec:add-exp});
    \item Industrial deployment in robotic warehouse grasping (Sec.~\ref{subsec:application});
    \item Implementation details of Any2Full (Sec.~\ref{supsec:details});
    \item Additional discussion (Sec.~\ref{sec:discussion}).
\end{itemize}

\section{Additional Experiments and Analyses}\label{sec:add-exp}
\subsection{Scale Consistency Analysis}\label{supsec:scale}

Monocular depth estimation (MDE) inherently suffers from spatial \textit{scale inconsistency}, where relative depth predictions exhibit spatially-varying scale factors when aligned to metric depth. 
To quantitatively measure this, we first align the predicted relative depth $\tilde{\hat{\mathbf{D}}}$ (in the disparity domain) to the ground-truth metric depth $\mathbf{D}$ via global least-squares fitting. Specifically, let $\mathbf{d} = 1/\mathbf{D}$ be the ground-truth disparity and $\tilde{\hat{\mathbf{D}}}$ be the predicted relative depth. We solve for a global scale $s$ and shift $t$ to minimize:
\begin{equation}
\min_{s, t} \| (s \cdot \tilde{\hat{\mathbf{D}}} + t) - \mathbf{d} \|^2.
\end{equation}
\vspace{-6pt}

The aligned metric depth prediction is then obtained as $\hat{\mathbf{D}} = 1 / (s \cdot \tilde{\hat{\mathbf{D}}} + t)$, where the reciprocal accounts for the disparity-to-depth conversion. 
We then partition the scene into non-overlapping regions $\{\mathcal{R}_k\}$ and compute the local scale factors:
\begin{equation}
s_k = \mathrm{median}_{i\in\mathcal{R}_k}(D_i / \hat{D}_i),
\end{equation}
where $D_i$ and $\hat{D}_i$ are the $i$-th pixels of $\mathbf{D}$ and $\hat{\mathbf{D}}$, respectively.
The variance of $\{s_k\}$ serves as a proxy for scale inconsistency and a perfectly consistent prediction would yield a uniform scale map where $s_k\!\approx\!1$ for all $k$, indicating that a single global transformation is sufficient for accurate relative-to-metric conversion.

\begin{figure}[h]
  \centering
  \includegraphics[width=0.63\linewidth]{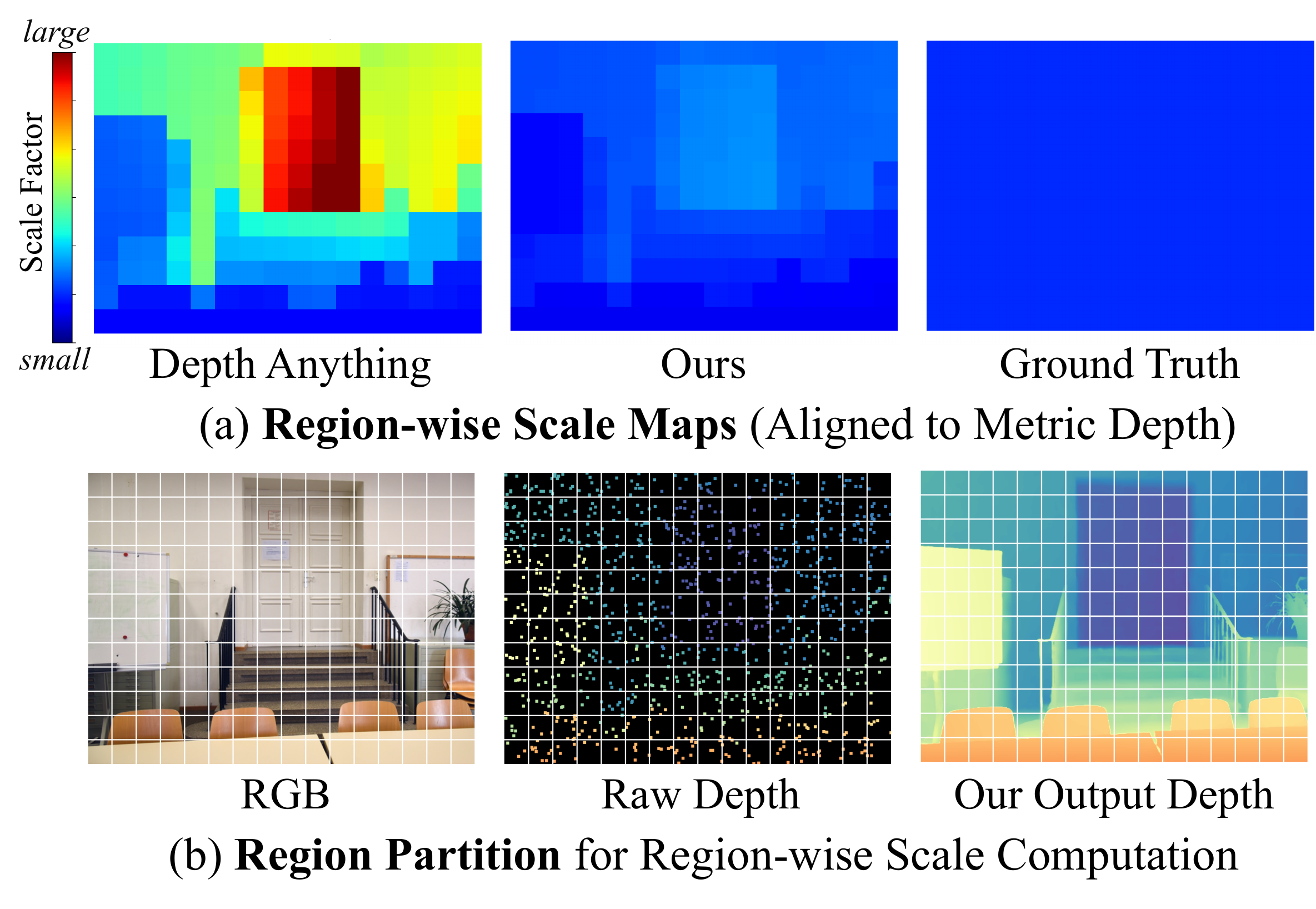}
  \vspace{-4pt}
  \caption{\textbf{Visualization of scale inconsistency.}
(a) The backbone MDE (Depth Anything) shows varying regional scale factors when aligned to metric depth (red = larger, blue = smaller), while our prediction yields nearly uniform scales consistent with the ground truth.
(b) The bottom row shows the region-partitioned RGB input, sparse depth, and our final result, illustrating how region-wise scales are computed.}

  \label{fig:scale_inconsistency}
\end{figure}

As visualized in Fig.~\ref{fig:scale_inconsistency}, the backbone MDE's scale map exhibits clear inter-region scale fluctuations, confirming its inherent scale inconsistency. 
In contrast, our scale-prompted model maintains a nearly uniform scale distribution across the entire scene. This demonstrates that Any2Full produces globally \textit{scale-consistent} predictions allowing for accurate metric recovery using only global alignment parameters.

\begin{figure}[!h]
  \centering
  \includegraphics[width=0.63\linewidth]{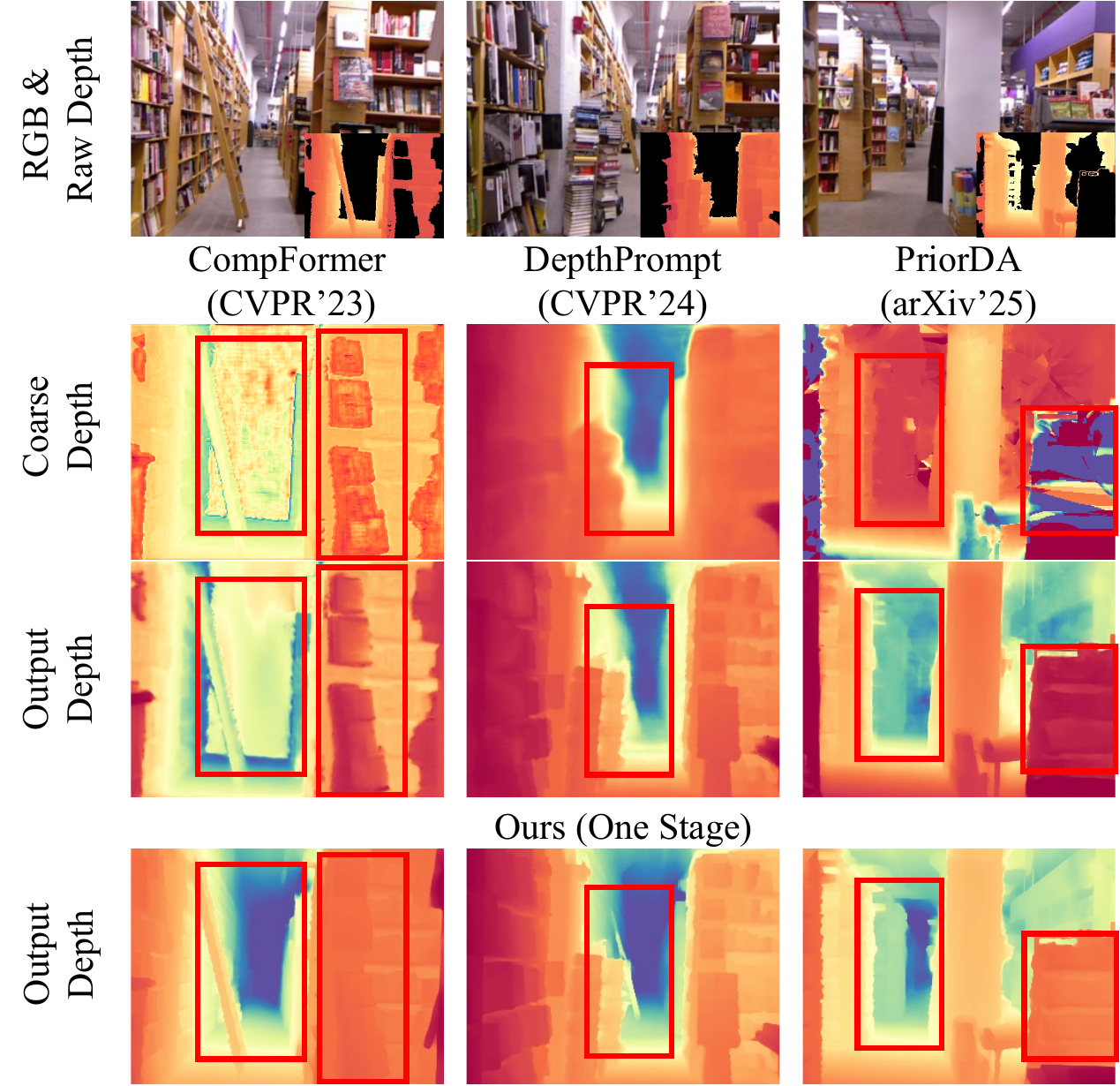}
  \vspace{-4pt}
  \caption{\textbf{Two-stage methods struggle to correct noisy coarse depths.}
We showcase the intermediate and final results on NYU-Depth V2 from two categories of two-stage pipelines: traditional RGBD fused methods (e.g., CompFormer~\cite{zhang2023completionformer}) and MDE-integrated methods (e.g., DepthPrompt~\cite{park2024depthprompt}, PriorDA~\cite{wang2025priorda}).
Both categories fail to correct noise and structural distortion in coarse depths, resulting in blurred details, disrupted structures, and visible artifacts in the final output.
In contrast, our Any2Full framework directly bridges sparse and dense domains in a single stage, achieving superior efficiency, accuracy, and robustness.
}
  \label{fig:2_stage}
  \vspace{-4pt}
\end{figure}

\subsection{Two-Stage Limitation Analysis}\label{supsec:2stage}
As mentioned in Fig.~\ref{fig:framework} of our main paper, previous two-stage approaches~\cite{zhang2023completionformer,park2020nonlocalspn,cheng2020cspn++,park2024depthprompt,wang2025priorda} 
first predict a coarse dense depth map, which often introduces noise and structural distortions. 
We provide additional visual comparisons of their intermediate and final results in Fig.~\ref{fig:2_stage}. 
Noisy coarse depths lead to blurred details, disrupted structures, and noticeable artifacts in the final predictions, 
revealing the limited robustness of such two-stage frameworks.

\subsection{Model Generalization}

\noindent\textbf{Additional Qualitative Results}

\begin{figure*}[ht]
\vspace{-10pt}
  \centering
  \includegraphics[width=0.8\textwidth]{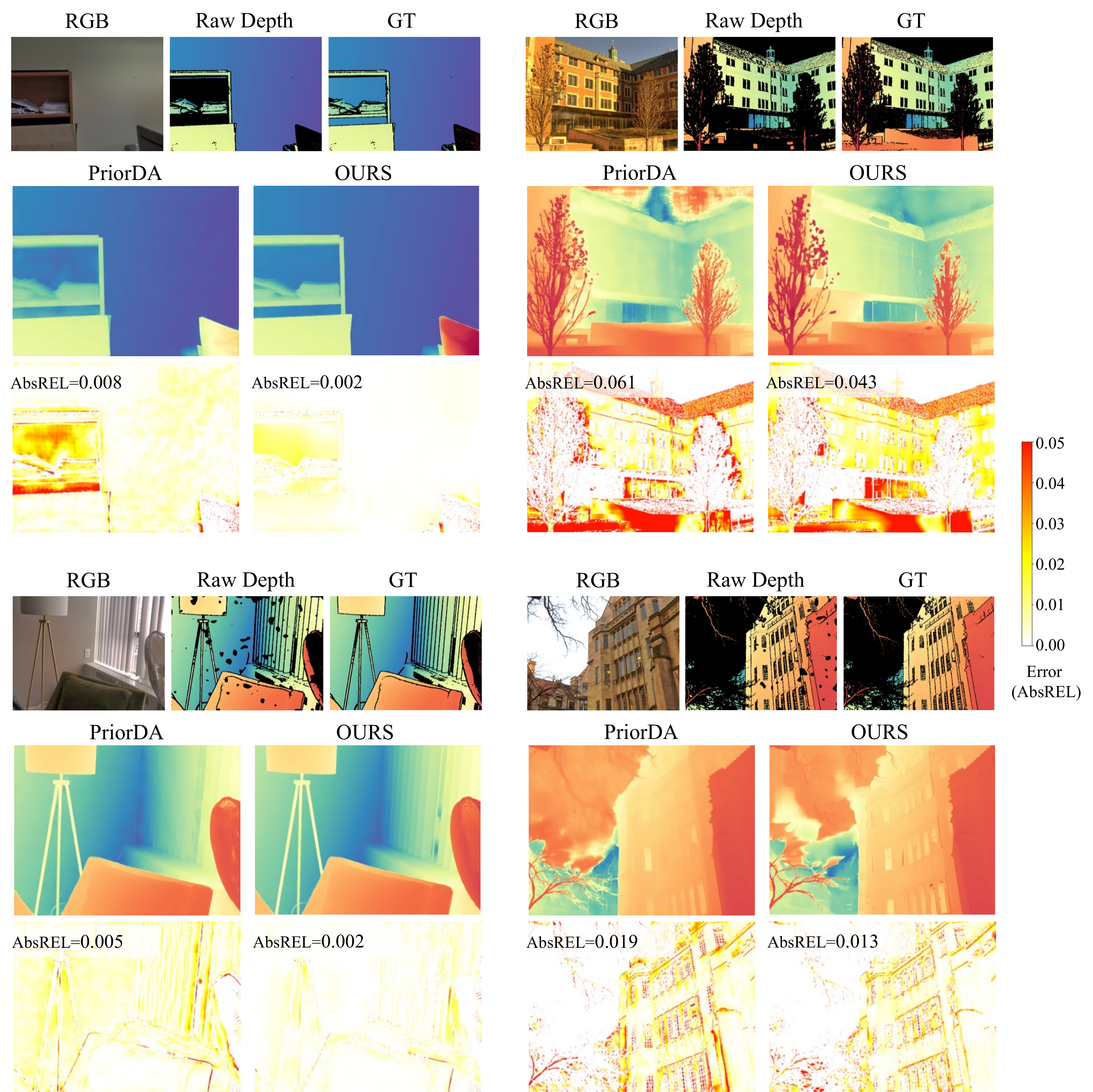}
  \vspace{-6pt}
  \caption{\textbf{Additional qualitative results on DIODE under different Hole patterns (semantic/small-block masking).} We compare PriorDA and our method (Any2Full) in predicted depth and error maps (AbsREL). 
Any2Full consistently preserves global geometry and achieves higher accuracy than PriorDA.}
  \label{fig:add-visiual-dh}
  \vspace{16pt}
\end{figure*}

\begin{figure*}[h]
\vspace{-6pt}
  \centering
  \includegraphics[width=0.8\textwidth]{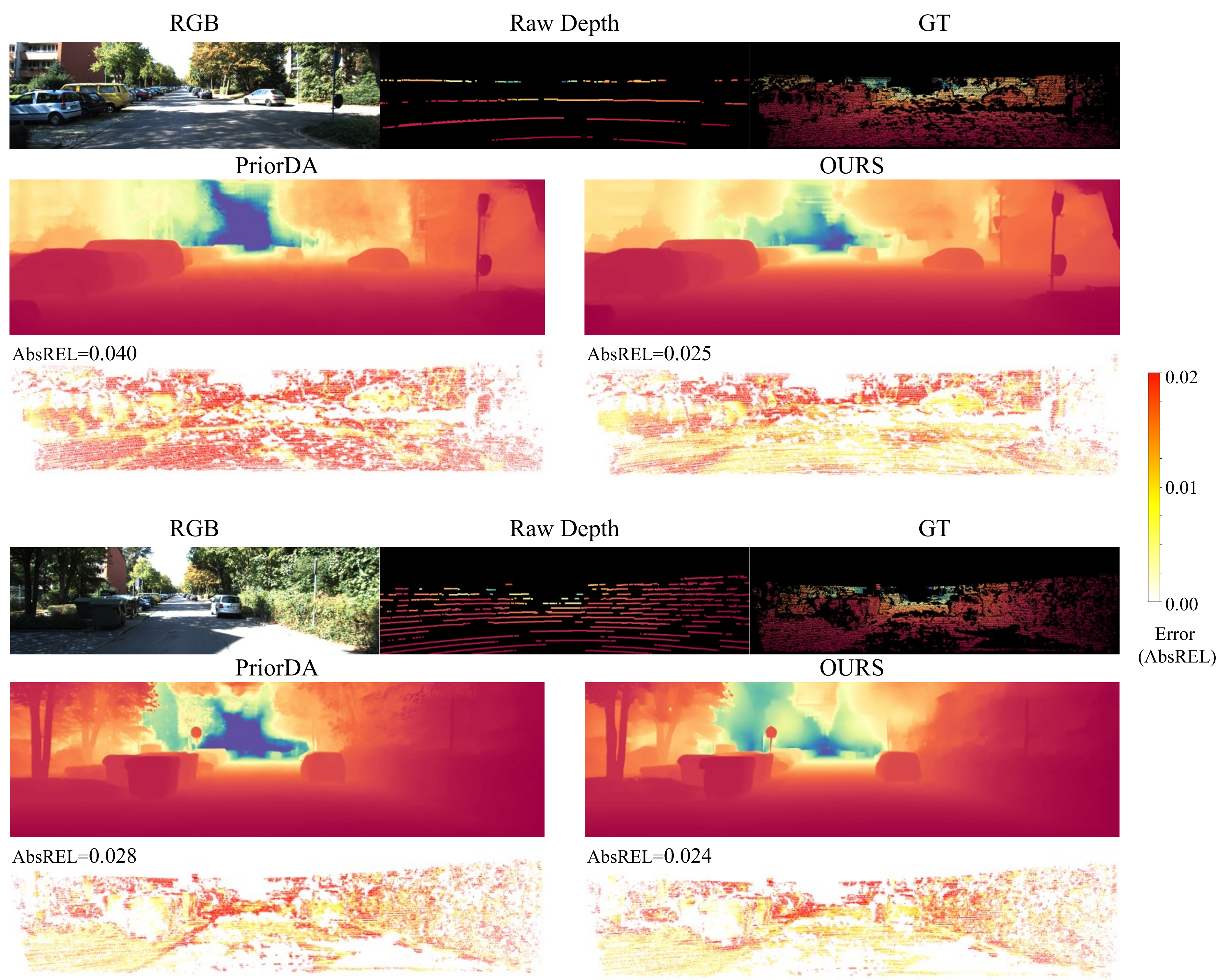}
  \vspace{-6pt}
  \caption{\textbf{Additional qualitative results on KITTI DC under different Sparse-LiDAR patterns (4L/16L).}}
  \label{fig:add-visiual-ks}
  \vspace{16pt}
\end{figure*}

\begin{figure*}[h]
\vspace*{-1.5\baselineskip}
  \centering
  \includegraphics[width=0.8\textwidth]{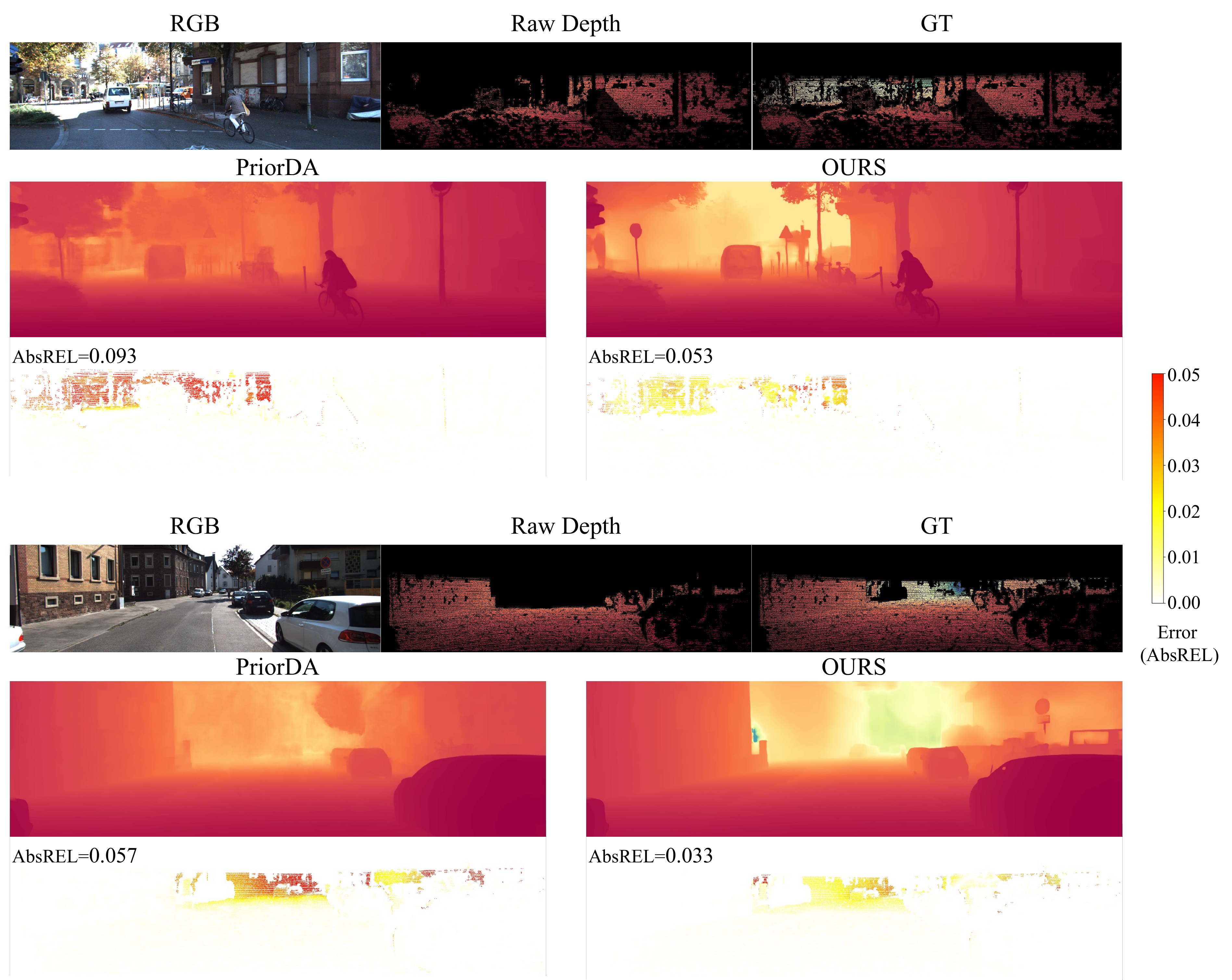}
  \vspace{-6pt}
  \caption{\textbf{Additional qualitative results on KITTI DC under the Range depth patterns. }}
  \label{fig:add-visiual-kr}
\end{figure*}

\begin{figure*}[h]
  \centering
  \includegraphics[width=0.8\linewidth]{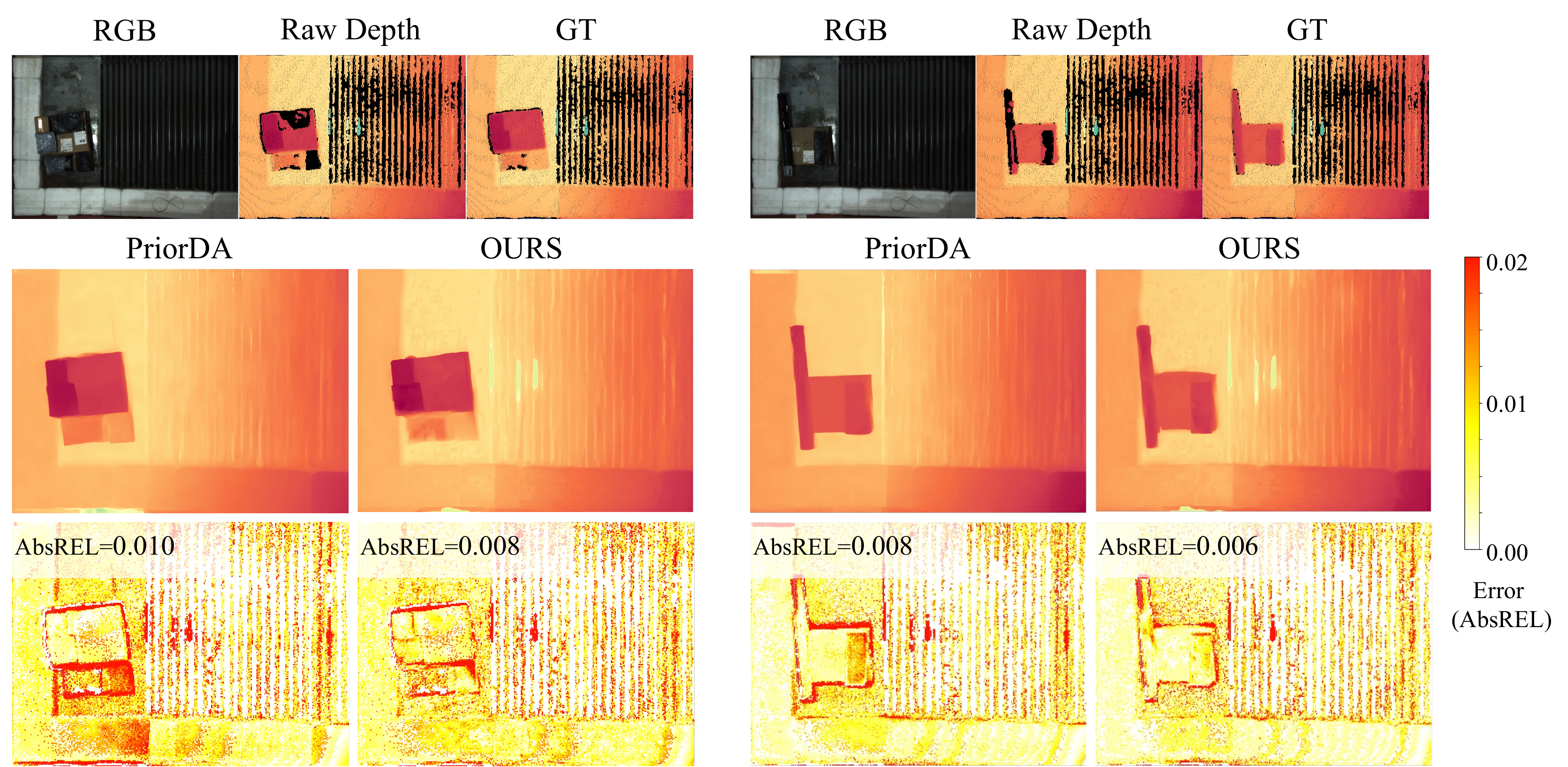}
  \caption{\textbf{Additional qualitative results on Logistic-Black under the Mixed depth pattern.}}
  \label{fig:add-visiual-jd}
\end{figure*}

\begin{figure*}[h]

  \centering
  \includegraphics[width=0.8\textwidth]{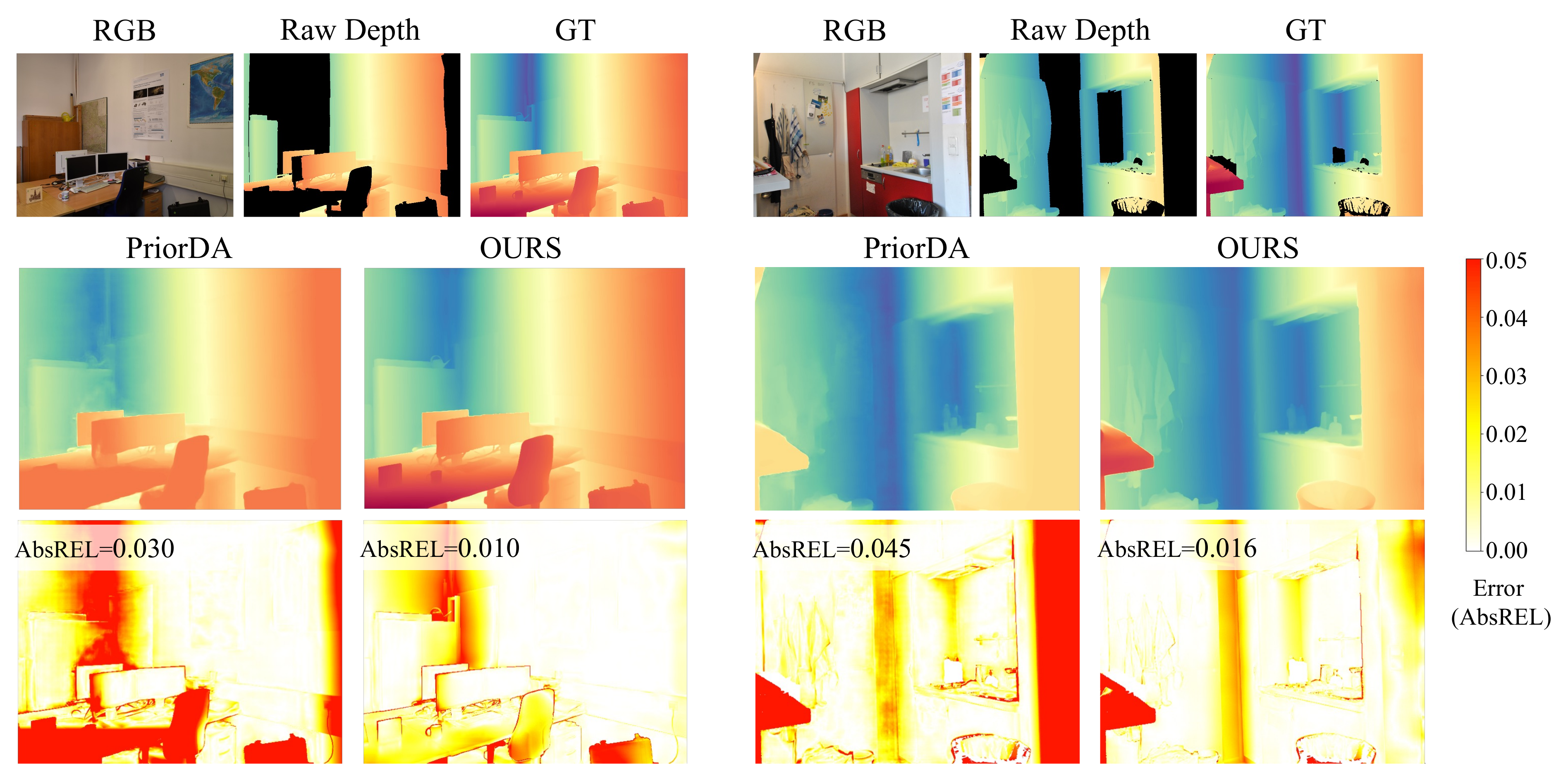}
  \caption{\textbf{Additional qualitative results on IBims-1 under the Range depth pattern.} }
  \label{fig:add-visiual-br}
\end{figure*}

\begin{figure*}[h]
  \centering
  \includegraphics[width=0.8\textwidth]{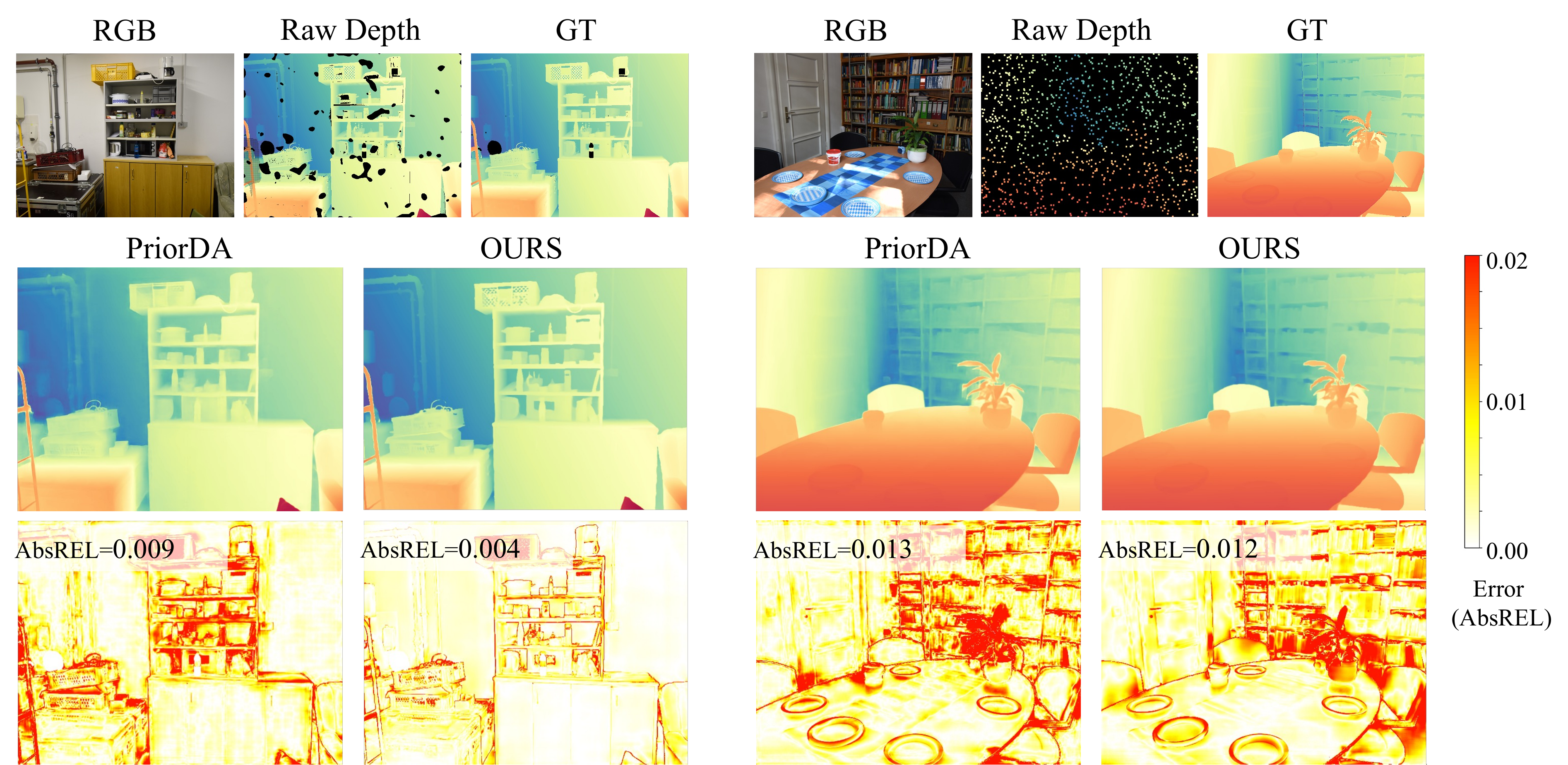}
  \caption{\textbf{Additional qualitative results on IBims-1 under the Hole and Sparse-Random depth patterns.}}
  \label{fig:add-visiual-bhr}
\end{figure*}

Due to space constraints in the main manuscript, we primarily present comparisons with the most representative patterns across several benchmarks. In this section, we provide extended qualitative results to further evaluate Any2Full across diverse domains and sparsity patterns.
We specifically choose PriorDA as our primary baseline for qualitative comparison because it stands as the most robust SOTA in terms of preserving fine geometric structures from MDE priors, as shown in Fig. \ref{fig:zero-dc-qual}.
As shown in the additional cases (Figs.~\ref{fig:add-visiual-dh},~\ref{fig:add-visiual-ks},~\ref{fig:add-visiual-kr},~\ref{fig:add-visiual-jd},~\ref{fig:add-visiual-br},~\ref{fig:add-visiual-bhr}), while PriorDA performs well, Any2Full further improves global scale consistency without the need for complex two-stage alignment.

\begin{figure}[t]
\vspace{-1pt}
    \centering
    \includegraphics[width=0.8\linewidth]{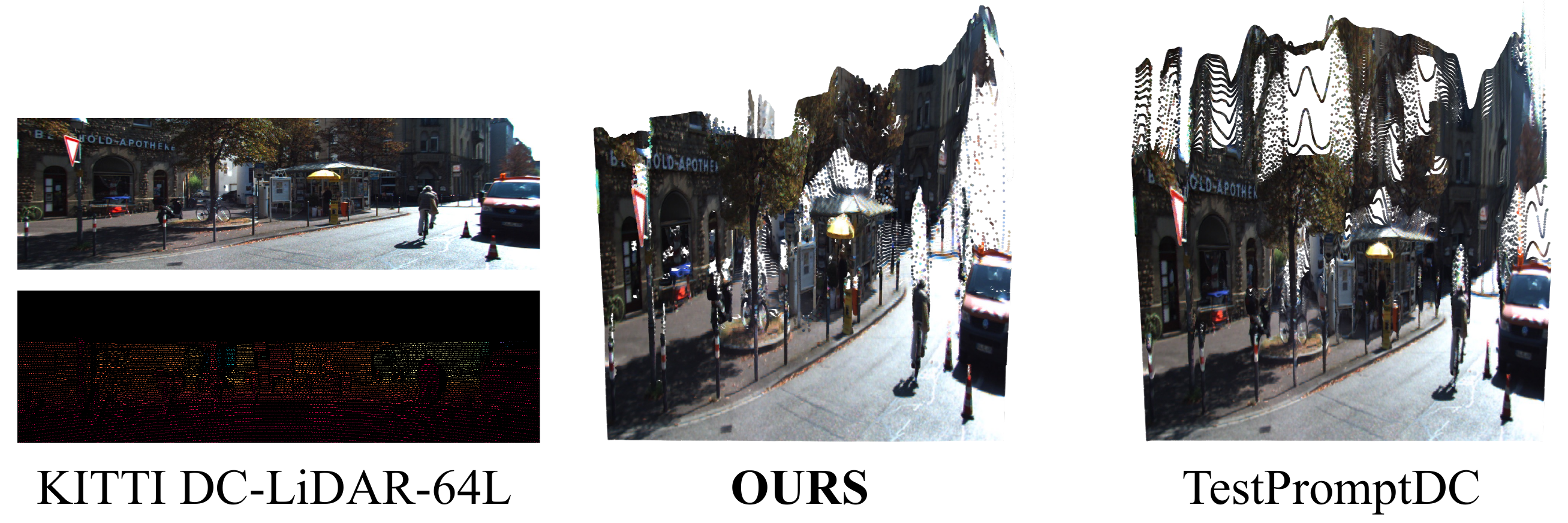}\\
    
    \includegraphics[width=0.8\linewidth]{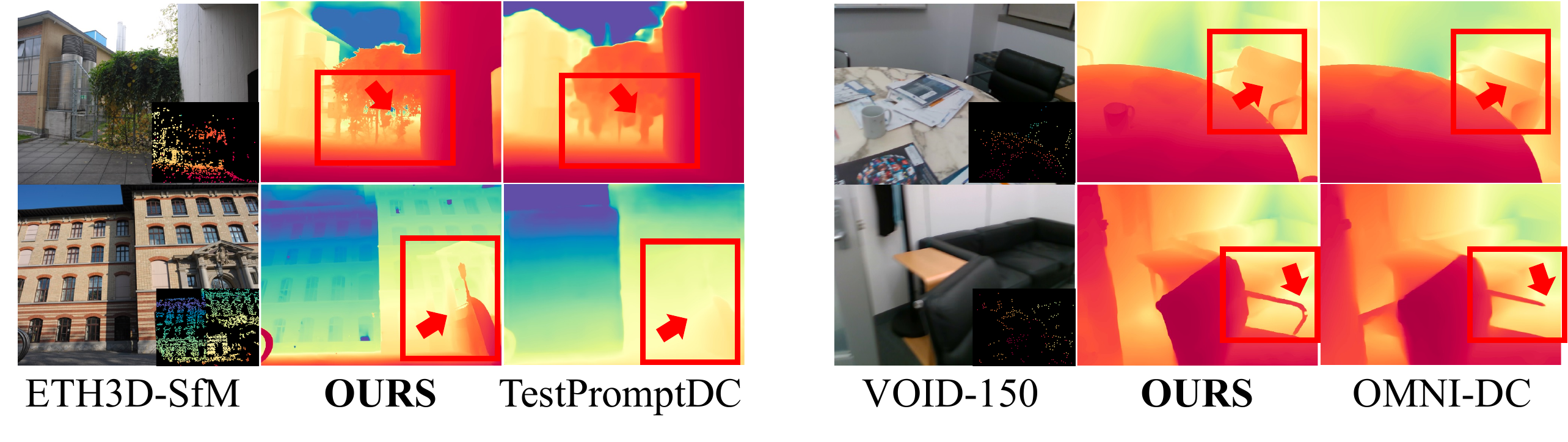}
    \vspace{-6pt}
    \caption{
    \textbf{Top:} 3D point-cloud reconstruction comparison. Any2Full yields fewer outliers and cleaner reconstructions than TestPromptDC.
    \textbf{Bottom:} Qualitative results on highly irregular sparse inputs, including ETH3D-SfM and VOID-150.
    }
    \vspace{-6pt}
    \label{fig:supp_geo_irregular}
\end{figure}

We further include geometric and irregular-sparsity visualizations in Fig.~\ref{fig:supp_geo_irregular}. The top row shows 3D point-cloud reconstruction comparisons, where Any2Full yields fewer outliers and cleaner reconstructions than TestPromptDC. The bottom row shows qualitative results on highly irregular sparse inputs, further supporting the pattern-agnostic generalization of Any2Full.

\begin{table}[h]
    \centering
    \caption{
    \textbf{Zero-shot depth super-resolution results.} 
Quantitative comparison under $\times 8$ downsampling setting. 
    }
     \vspace{-6pt}
    \resizebox{0.7\linewidth}{!}{
    \begin{tabular}{l *{3}{cc}}
        \toprule
        & \multicolumn{2}{c}{\textbf{NYU-Depth V2}} &
          \multicolumn{2}{c}{\textbf{IBims-1}} &
          \multicolumn{2}{c}{\textbf{KITTI DC}} \\
        \cmidrule(lr){2-3} \cmidrule(lr){4-5} \cmidrule(lr){6-7}
        \multicolumn{1}{c}{Method} &
        \multicolumn{1}{c}{AbsREL} & \multicolumn{1}{c}{RMSE} &
        \multicolumn{1}{c}{AbsREL} & \multicolumn{1}{c}{RMSE} &
        \multicolumn{1}{c}{AbsREL} & \multicolumn{1}{c}{RMSE} \\
        \midrule

        \makecell[l]{Depth Anything{+}LS } &
        0.061 & 0.302 &
        0.038 & 0.346 &
        0.078 & 3.534 \\

        \makecell[l]{Marigold{+}LS } &
        0.080 & 0.319 &
        0.058 & 0.349 &
        0.198 & 6.044 \\
        \midrule

        \makecell[l]{PromptDA} &
        0.030 & 0.221 &
        0.016 & 0.158 &
        0.036 & 2.745 \\

        \makecell[l]{Marigold-DC} &
        0.018 & 0.127 &
        0.014 & 0.174 &
        0.050 & 2.442 \\

        \makecell[l]{PriorDA } &
        \best{0.012} & \best{0.095} &
        \second{0.011} & \best{0.138} &
        \second{0.026} & \second{1.952} \\
        \midrule

        Ours &
        \second{0.012} & \second{0.106} &
        \best{0.010} & \second{0.142} &
        \best{0.020} & \best{1.603} \\
        \bottomrule
    \end{tabular}}
    \vspace{-10pt}
    \label{tab:depth_super_resolution}
\end{table}

\noindent\textbf{Task Generalization} 
To further assess the robustness of our model under different setups, 
we also evaluate Any2Full on zero-shot depth super-resolution as shown in~\cref{tab:depth_super_resolution}. 
In this setting, low-resolution raw depth is obtained by downsampling the ground-truth depth by $\times 8$ 
using bilinear resizing. 
Unlike irregular sparse patterns, this task introduces low-frequency, grid-like degradation. 
Moreover, downsampling reduces the reliability of individual depth value, on which our method relies to anchor scale.
Despite these differences, Any2Full still performs competitively, showing that
our approach retains generalization ability beyond its primary sparse depth regime.

\subsection{Ablation Studies}\label{supsec:ablation}

\noindent\textbf{Module Ablation.}
Tab.~\ref{tab:ablation} in the main paper presents a module ablation study of our model. 
The results show that the global propagation module significantly improves performance, 
especially under range and sparse depth patterns.

\begin{figure}[t]
\vspace{-4pt}
  \centering
  \includegraphics[width=0.67\linewidth]{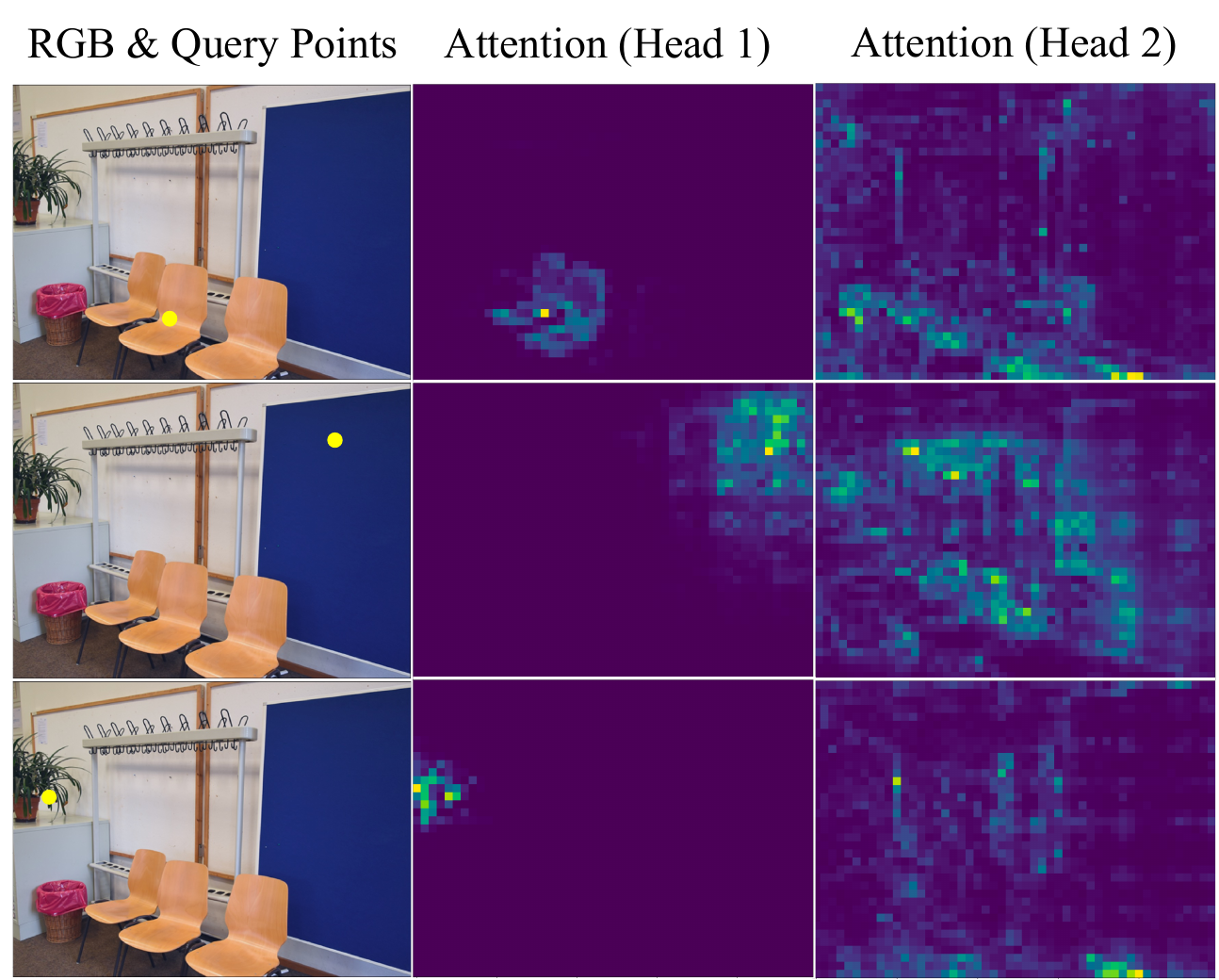}

  \caption{\textbf{Visualization of attention propagation} from different query points (yellow dots). 
Each row corresponds to a distinct query location shown in the left column (RGB \& query points). 
The middle and right columns visualize attention maps from two representative heads of the global propagation module in layer~2, 
illustrating how the propagation extends from nearby to distant regions. 
Head~1 focuses on refining nearby patches with similar geometry, 
whereas Head~2 captures long-range scale consistency. 
Together, they achieve a balanced propagation of scale cues, 
enhancing both \textbf{local scale recovery} and \textbf{global scale consistency}.}
  \label{fig:att}
  \vspace{-4pt}
\end{figure}

To further understand its behavior, we visualize the global propagation attention maps in Fig.~\ref{fig:att}, revealing distinct propagation patterns from nearby to distant regions. 
Such complementary propagation enables Any2Full to diffuse scale cues more effectively across spatially irregular and sparse inputs.

Notably, the attention weights in our global propagation module are computed solely from MDE features, 
ensuring that the propagation is guided by geometric structures rather than sparse depth patterns. 
This design preserves robustness across varying depth sampling distributions. 
Tab.~\ref{tab:cross-attn} shows that replacing our MDE-guided propagation with standard 
cross-attention leads to noticeable performance drops.

\begin{table}[t]
\centering
\caption{\textbf{Comparison with cross-attention-based propagation.} 
We replace our MDE-guided propagation with standard cross-attention 
to test robustness under different depth patterns.
Our method consistently achieves lower AbsREL across all settings.}

\scriptsize
\setlength{\tabcolsep}{3pt}
\begin{tabular}{l ccc ccc}
\toprule
& \multicolumn{3}{c}{\textbf{IBims-1 (AbsREL)}} & \multicolumn{3}{c}{\textbf{KITTI DC (AbsREL)}} \\
\cmidrule(lr){2-4} \cmidrule(lr){5-7}
\textbf{Method} & Hole & Range & Sparse & Hole & Range & Sparse \\
\midrule
Cross-attention & 0.009 & 0.057 & 0.016 & 0.015 & 0.057 & 0.029 \\
Ours & \best{0.007} & \best{0.030} & \best{0.013} & \best{0.011} & \best{0.046} & \best{0.021} \\
\bottomrule
\end{tabular}
\label{tab:cross-attn}
\vspace{-6pt}
\end{table}

\vspace{10pt}
\noindent\textbf{Loss Ablation.}
\begin{table}[!h]
\centering
\caption{
\textbf{Loss ablation} of Any2Full based on DA-B.
Each loss term is incrementally added:
$\mathcal{L}_{\text{ssi}}$ (global alignment), $\mathcal{L}_{\text{gm}}$ (fine geometry),
$\mathcal{L}_{\text{r-ssim}}$ (relative structure), and $\mathcal{L}_{\text{anchor}}$ (sparse anchoring).
}

\scriptsize
\resizebox{0.8\linewidth}{!}{
\begin{tabular}{cccc ccc ccc}
\toprule
\textbf{$\mathcal{L}_{\text{ssi}}$} & \textbf{$\mathcal{L}_{\text{gm}}$} & \textbf{$\mathcal{L}_{\text{r-ssim}}$} & \textbf{$\mathcal{L}_{\text{anchor}}$} &
\multicolumn{3}{c}{\textbf{IBims-1 (AbsREL)}} &
\multicolumn{3}{c}{\textbf{KITTI DC (AbsREL)}} \\
\cmidrule(lr){5-7} \cmidrule(lr){8-10}
& & & & Hole & Range & Sparse & Hole & Range & Sparse \\
\midrule
\checkmark &  &  &  & 0.0086 & 0.0336 & 0.0149 & 0.0132 & 0.0661 & \underline{0.0189} \\
\checkmark & \checkmark &  &  & 0.0083 & \underline{0.0338} & 0.0148 & \underline{0.0132} & \underline{0.0600} & 0.0191 \\
\checkmark & \checkmark & \checkmark &  & \underline{0.0083} & 0.0348 & \underline{0.0148} & 0.0136 & 0.0632 & 0.0192 \\
\checkmark & \checkmark & \checkmark & \checkmark & \textbf{0.0077} & \textbf{0.0306} & \textbf{0.0131} & \textbf{0.0115} & \textbf{0.0587} & \textbf{0.0173} \\
\bottomrule
\end{tabular}}
\vspace{-6pt}
\label{tab:loss-ablation}
\end{table}

Tab.~\ref{tab:loss-ablation} reports the ablation of different loss terms, 
corresponding to the formulations in Sec.~\ref{supsec:loss}. 
Both $\mathcal{L}_{\text{gm}}$ for fine-geometry supervision and 
$\mathcal{L}_{\text{anchor}}$ for sparse-depth anchoring consistently improve performance. 
While adding $\mathcal{L}_{\text{r-ssim}}$ encourages locally smooth depth predictions, which may slightly constrain the model’s ability to expand depth estimates beyond limited observations in Range pattern. We retain $\mathcal{L}_{\text{r-ssim}}$ in our final objective to prioritize structural consistency, which are essential for stable performance in downstream applications.

\section{Real-World Industrial Deployment}\label{subsec:application}
\vspace{-6pt}
In industrial robotic warehouse grasping, precise depth sensing is essential for 3D geometric perception of incoming packages and the subsequent computation of grasp poses for robotic manipulation. However, industrial environments are plagued by the "black package problem"—items with high-absorption surfaces that constitute approximately 0.5\% of the millions of daily warehouse items. While statistically a minority, these black packages pose a significant challenge for active depth sensors, such as ToF or structured light cameras, as the emitted light signals are largely absorbed by the dark surfaces rather than reflected back to the sensor~\cite{wang2022depthhole}. This physical limitation leads to severe depth loss, preventing accurate perception of package geometry. Consequently, the resulting grasp failures necessitate costly manual intervention, while package deformation due to suboptimal grasp points directly compromises the customer experience and logistics service reliability.

Addressing this through supervised learning is bottlenecked by the extreme difficulty of acquiring large scale, high-fidelity, ground-truth RGB-D data for black packages. This necessitates a domain-general and pattern-agnostic depth completion solution. After validating the efficiency and effectiveness of our Any2-Full on the Logistic-Black dataset (see \cref{sec:experiment}), we transition from offline benchmarks to a live production-grade robotic cell to evaluate its practical utility and, ultimately, achieve sustained long-term deployment. Given the strict real-time constraints and the engineering complexity of the industrial environment, we focus on verifying the practical reliability of Any2Full in resolving the sensor's fundamental failure, rather than a comparative study of different algorithms.

\begin{figure}[t]
    \centering
    \includegraphics[width=1\linewidth]{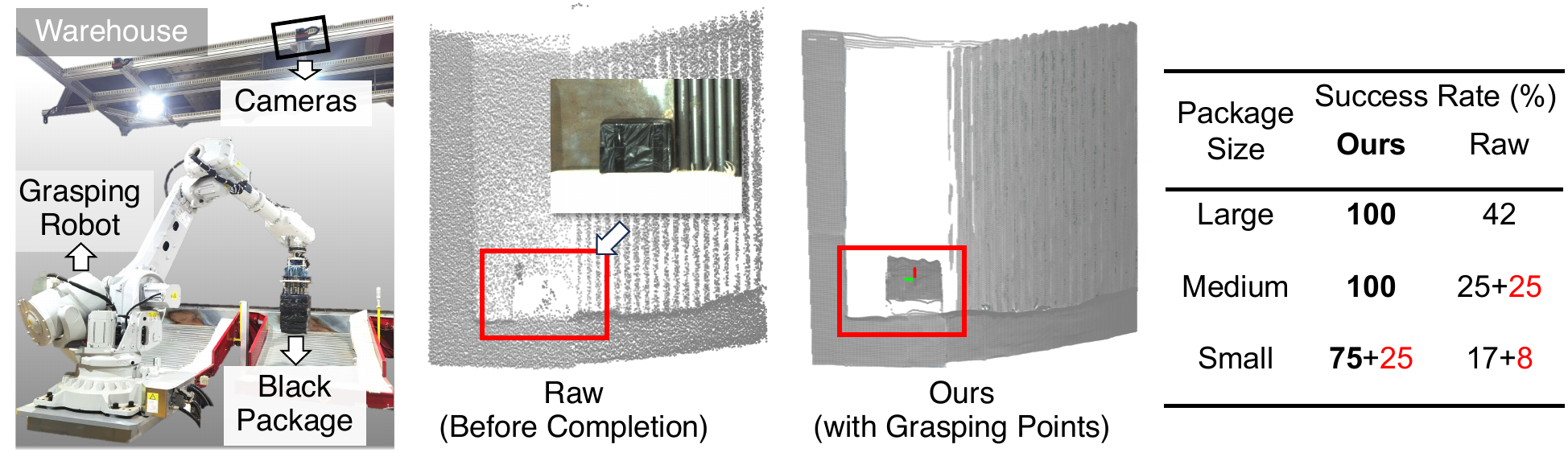}
    \caption{\textbf{Robotic warehouse grasping setup and depth results.} The table reports grasping success rates across package sizes. Black indicates successful grasps without damage, while red denotes successful but deformed packages due to poor grasp points. }
    \label{fig:warehouse}

\end{figure}

\vspace{4pt}
\noindent\textbf{Deployment Setup.}
Fig.~\ref{fig:warehouse} (left) shows our real-world robotic warehouse setting. The system employs ABB IRB~6700-150/3.20 manipulators equipped with Vzense DS77C~Pro ToF cameras (640$\times$480 depth resolution) and MV-CS050-10GC RGB cameras (2448$\times$2048 resolution). Grasp poses are computed using a RANSAC-based~\cite{fischler1981random} algorithm. Our model Any2Full is deployed on an NVIDIA GeForce RTX 4080 Super GPU, achieving an inference rate of 10 Hz (0.1 s per frame), which satisfies the real-time constraints of industrial cycle times.

\begin{figure}[t]
  \centering
  \includegraphics[width=0.56\linewidth]{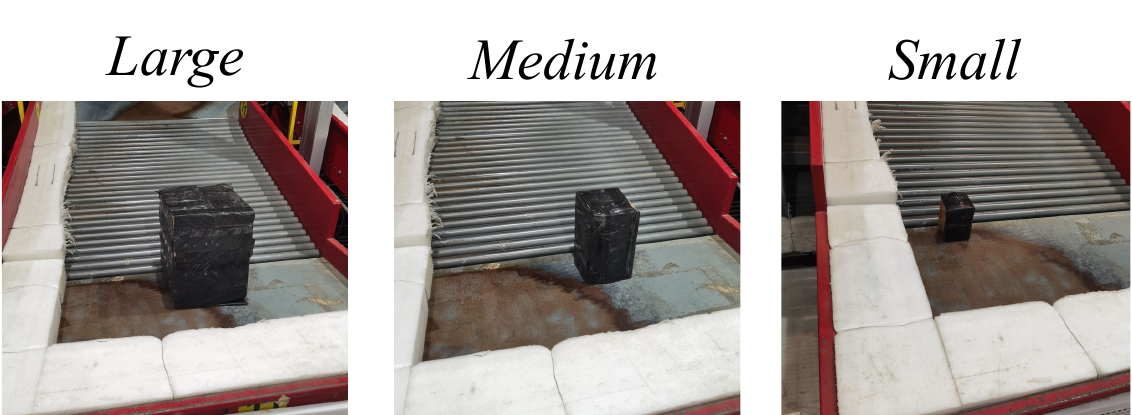}
  \caption{\textbf{Examples of black packages used in the grasping evaluation.}
From left to right: \textit{large} (smallest edge $>$ 30\,cm), 
\textit{medium} (10–30\,cm), 
and \textit{small} ($<$ 10\,cm). }
  \label{fig:packages}
  \vspace{-6pt}
\end{figure}

\vspace{4pt}
\noindent\textbf{Grasping Evaluation.}
We conducted comparative grasping experiments between a baseline system using raw depth data and a system enhanced by our depth completion model.  Trials focused on black packages of three scales: \textit{Large}, \textit{Medium}, and \textit{Small}, as shown in Fig.~\ref{fig:packages}. For each size category, we performed 12 paired trials under identical conditions 
(i.e., same package, position and lighting) to ensure an unbiased comparison.

As quantified in the table in Fig.~\ref{fig:warehouse}, raw depth sensing yields a grasp success rate of only 39.0\%, with 33.3\% of those successes resulting in package deformation due to imprecise geometry. Small packages are particularly challenging because their limited reflective area causes less reliable ToF depth, leading to increased grasp failure rates and package deformation. By integrating Any2Full, the robot reconstructs the complete 3D geometry of these high-absorption items, increasing the grasp success rate to 100\% and ensuring 91.6\% of grasps are performed without any package deformation.

\vspace{4pt}
\noindent\textbf{Long-Term Deployment Stability.}
Once verified that our model significantly improves grasp success rates within the robotic cell, we leveraged its low inference overhead to perform a seamless integration. Our model was deployed in an active cell within a warehouse belonging to one of the world’s largest logistics providers~\cite{xinyue,zejun,zhiyuan}. During a continuous one-week operational trial, the system maintained a stable grasp success rate of 98.86\% on standard packages. This sustained performance demonstrates that our dense depth predictions introduce no additional geometric noise or artifacts and remain robust under continuous real-world operational conditions, proving the model's readiness for large-scale industrial use.

\section{Implementation Details of Any2Full}\label{supsec:details}
This section provides additional implementation details of Any2Full, including the normalization of sparse depth inputs, 
the embedding of sparsity masks, and the training losses used for supervision.

\subsection{Normalization of Sparse Depth}\label{supsec:norm}
To enable scale-consistent supervision, all input and predicted depths are converted into the 
\textit{relative depths} adopted by MDE models, 
which corresponds to the disparity space in Depth Anything~v2. 
Given the sparse metric depth $\mathbf{D}_s$, we first convert its valid pixels to the disparity (inverse depth) domain:
\begin{equation}
d_i = 1 / D_{s,i},
\end{equation}
where $d_i$ denotes the disparity value at pixel $i$. We then normalize these values to remove global scale and shift:
\begin{equation}
\tilde{d}_i = \frac{d_i - \mu(d)}{\sigma(d)},
\end{equation}
where $\mu(d)$ and $\sigma(d)$ denote the mean and standard deviation calculated over all valid $d_i$. The resulting normalized elements $\tilde{d}_i$ form the normalized sparse depth input $\tilde{\mathbf{D}}_s$ used in our model.

For symbol simplicity, we use the tilde symbol $(\tilde{\cdot})$ to denote normalized values in the disparity domain.
All supervised training losses are computed on relative depth $(\tilde{\mathbf{D}},\,\tilde{\hat{\mathbf{D}}})$, 
while evaluation metrics are measured in the metric depth $(\mathbf{D},\,\hat{\mathbf{D}})$.

\subsection{Sparse Depth Embedding}\label{supsec:embed}
We design a mask-aware sparse depth embedding to distill scale information from sparse measurements 
and encode them as patch-level depth features for subsequent local enrichment module.  
Given the sparse depth map $\tilde{\mathbf{D}}_s$, we first patchify it into non-overlapping regions 
and construct a set of multi-size depth maps within each patch 
by progressively downsampling the sparse depth to different spatial resolutions, enabling coarse-to-fine scale extraction.
For each size depth map, invalid pixels are filled by nearest-value interpolation to obtain a dense depth map.  
Each filled depth map, together with its corresponding validity mask, is then fed into a linear layer to jointly encode scale cues and confidence.
The averaged feature across multi-size maps forms the patch embedding.  
A learnable positional embedding is added, and the mean of all patch embeddings is used as a CLS token
to align with the representation format of MDE encoders, producing the final patch-level depth features $\mathbf{F}_{dep}$.

In addition, a binary patch mask is recorded, 
where a value of~0 indicates patches with no valid sparse measurements.  
This mask is passed to the global propagation module regulating cross-region attention.

\subsection{Loss Definition} \label{supsec:loss}
We train Any2Full using a combination of loss terms that jointly supervise 
global alignment, structural accuracy, and sparse anchoring. For consistent notation with the main text, let $\tilde{D}_i$ and $\tilde{\hat{D}}_i$ denote the normalized sparse input and predicted relative values in the disparity domain, respectively.
Let $\Omega$ denote all valid supervision pixels and $\Omega_s \subset \Omega$ the sparse anchors.  

\noindent \textbf{Scale- and Shift-Invariant Loss}~\cite{yang2024depthanythingv2}.
To supervise global relative alignment, we adopt an $\ell_1$ loss:
\begin{equation}
\mathcal{L}_{\text{ssi}} =
\frac{1}{|\Omega|}\sum_{i \in \Omega}
|\tilde{\hat{D}}_i - \tilde{D}_i|.
\end{equation}

\noindent \textbf{Gradient Matching Loss}~\cite{yang2024depthanythingv2}.
To preserve geometric edges and fine details:
\begin{equation}
\mathcal{L}_{\text{gm}} =
\frac{1}{|\Omega|}\sum_{i \in \Omega}
(|\nabla_x \tilde{\hat{D}}_i - \nabla_x \tilde{D}_i| +
 |\nabla_y \tilde{\hat{D}}_i - \nabla_y \tilde{D}_i|).
\end{equation}

\noindent \textbf{Sparse Anchor Consistency}.
To maintain consistency with sparse observations by applying the same $\mathcal{L}_{\text{ssi}}$ on sparse anchors:
\vspace{-4pt}
\begin{equation}
\mathcal{L}_{\text{anchor}} =
\frac{1}{|\Omega_s|}\sum_{i \in \Omega_s}
|\tilde{\hat{D}}_i - \tilde{D}_i|.
\end{equation}
\noindent \textbf{Relative Structure Similarity} (Optional).
Following~\cite{hyoseok2025zero}, we optionally include:
\begin{equation}
\mathcal{L}_{\text{r-ssim}} = \frac{1}{|\Omega|}\sum_{i \in \Omega} \left( 1 - \text{SSIM}(\tilde{D}_i, \tilde{\hat{D}}_i) \right),
\end{equation}
where $\text{SSIM}(\cdot)$ computes the local structural similarity index between  $\tilde{D}$ and  $\tilde{\hat{D}}$ within a $3 \times 3$ window centered at pixel $i$.

\noindent The overall objective is:
\begin{equation}
\mathcal{L}_{\text{total}} =
\lambda_1\mathcal{L}_{\text{ssi}} +
\lambda_2\mathcal{L}_{\text{gm}} +
\lambda_3\mathcal{L}_{\text{anchor}} +
\lambda_4\mathcal{L}_{\text{r-ssim}},
\end{equation}
with $\lambda_1=\lambda_2=\lambda_3=\lambda_4=0.5$ unless otherwise specified.

\subsection{Evaluation Metrics}\label{subsec:metrics}
To quantitatively evaluate the model performance in metric space, we adopt two standard metrics computed on the predicted metric depth $\hat{D}_i$ and ground-truth depth $D_i$:

\paragraph{Absolute Relative Error (AbsREL).}
\begin{equation}
\text{AbsREL} = \frac{1}{|\Omega|} \sum_{i \in \Omega} 
\frac{|\hat{D}_i - D_i|}{D_i}.
\end{equation}

\paragraph{Root Mean Squared Error (RMSE)}, reported in meters.
\begin{equation}
\text{RMSE} = \sqrt{ \frac{1}{|\Omega|} 
\sum_{i \in \Omega} (\hat{D}_i - D_i)^2 }.
\end{equation}
Both metrics are computed over all valid pixels in $\Omega$.

\section{Discussion}\label{sec:discussion}
\vspace{-2pt}
Our method benefits from the strong geometric priors provided by monocular depth estimation backbones. 
Recently, Depth Anything~\cite{yang2024depthanythingv1,yang2024depthanythingv2} released v3, which extends the framework to arbitrary visual inputs and provides a new metric-depth prediction model. 
Nevertheless, their results indicate that monocular depth estimation alone still struggles to recover reliable absolute scale 
without metric measurements. 

We believe our formulation remains complementary to these advancements. 
In future work, integrating Any2Full with Depth Anything v3 may further improve monocular depth completion 
and potentially extend the framework to arbitrary-view depth completion for more general setups.

\vspace{2pt}

%

\end{document}